\documentclass[letterpaper,10pt,conference]{ieeeconf}
\IEEEoverridecommandlockouts
\usepackage[T1]{fontenc}
\usepackage[utf8]{inputenc}
\usepackage{amsmath,amssymb,amsthm}
\usepackage{graphicx}
\graphicspath{{figs/}}
\usepackage{cite}
\usepackage[hyphens]{url}
\usepackage{xcolor}
\usepackage{tikz}
\usetikzlibrary{arrows.meta,positioning,calc}
\usepackage{booktabs}
\usepackage{array}
\usepackage{microtype}
\usepackage{pgfplots}\pgfplotsset{compat=1.18}\usepgfplotslibrary{fillbetween}

\theoremstyle{plain}
\newtheorem{theorem}{Theorem}
\newtheorem{lemma}{Lemma}
\newtheorem{proposition}{Proposition}
\newtheorem{corollary}{Corollary}
\theoremstyle{definition}
\newtheorem{definition}{Definition}
\newtheorem{example}{Example}
\newtheorem{remark}{Remark}
\newtheorem*{assumptionAbase}{Assumption A}
\newtheorem*{assumptionBbase}{Assumption B}
\makeatletter
\newenvironment{assumptionA}[1]{\def\@currentlabel{A}\begin{assumptionAbase}[#1]}{\end{assumptionAbase}}
\newenvironment{assumptionB}[1]{\def\@currentlabel{B}\begin{assumptionBbase}[#1]}{\end{assumptionBbase}}
\makeatother

\newcommand{\N}{\mathcal{N}}
\newcommand{\Q}{\mathcal{Q}}
\newcommand{\B}{\mathcal{B}}
\newcommand{\F}{\mathcal{F}}
\newcommand{\X}{\mathcal{X}}
\newcommand{\R}{\mathcal{R}}
\newcommand{\Tc}{T_c}
\newcommand{\Tu}{T_u}
\newcommand{\En}{\mathrm{En}}
\newcommand{\Enc}{\mathrm{En}_c}
\newcommand{\Enu}{\mathrm{En}_u}
\newcommand{\Kstar}{K^{\star}}
\newcommand{\Kcirc}{K^{\uparrow}}
\newcommand{\Sstar}{S^{\star}}
\newcommand{\SK}{S_{\Kcirc}}
\newcommand{\Sopt}{S_{J_{\min}}}
\newcommand{\Jopt}{J_{\min}}
\newcommand{\Sclass}{\mathcal{S}}
\newcommand{\ca}{c_{a}}
\newcommand{\Pre}{\mathrm{Pre}}
\newcommand{\Post}{\mathrm{Post}}
\newcommand{\UPre}{\mathrm{UPre}}
\newcommand{\pl}[1]{p_{\mathrm{#1}}}
\newcommand{\tr}[1]{t_{\mathrm{#1}}}
\renewcommand{\bot}{\mathord{\perp}}
\newcommand{\E}{\mathbb{E}}

\makeatletter
\def\bstctlcite{\@ifnextchar[{\@bstctlcite}{\@bstctlcite[@auxout]}}
\def\@bstctlcite[#1]#2{\@bsphack\@for\@citeb:=#2\do{\edef\@citeb{\expandafter\@firstofone\@citeb}\if@filesw\immediate\write\csname #1\endcsname{\string\citation{\@citeb}}\fi}\@esphack}
\makeatother
\definecolor{RiskRed}{RGB}{175,40,40}% figures: refusal (same RGB as figs/*.pdf)
\colorlet{MonFill}{black!14}% figures: monitor-place fill (same grey as figs/*.pdf)

\title{\LARGE \bf
Certified Execution Harness for Agentic AI: Safe and Least-Intrusive Petri-Net Supervisory Control Between Reasoning and Action}

\author{Chao Gu$^{1}$ and Quanyan Zhu$^{2}$
\thanks{$^{1}$C.~Gu is with the School of Electronics, Electrical Engineering and Computer Science, Queen's University Belfast, Belfast, UK {\tt\small cgu@qub.ac.uk; chaogu.ac@gmail.com}}\thanks{$^{2}$Q.~Zhu is with the Department of Electrical and Computer Engineering, New York University, Brooklyn, NY 11201, USA {\tt\small qz494@nyu.edu}}
}

\begin{document}
% Layout: float spacing slightly tighter than the ieeeconf defaults (1.7\baselineskip and 0.5\baselineskip),
% which keeps the paper at 17 pages without changing any content.
\setlength{\textfloatsep}{1.2\baselineskip plus 0.2\baselineskip minus 0.4\baselineskip}
\setlength{\dbltextfloatsep}{1.2\baselineskip plus 0.2\baselineskip minus 0.4\baselineskip}
\setlength{\abovecaptionskip}{5pt}

\bstctlcite{IEEEexample:BSTcontrol}% print repeated author names in full

\maketitle
\thispagestyle{plain}
\pagestyle{plain}

\renewcommand{\thefootnote}{\fnsymbol{footnote}}
\begin{abstract}
This paper studies the problem of safe tool execution for large language model (LLM) agents under prompt injection. Using Petri nets, we formalize the \emph{execution harness} between the LLM and its tools as a supervisory control problem, in which \emph{marks} represent capabilities, approvals and budgets. Since a transition fires only if the marks it needs are present, an injected request whose preconditions are unmet or already spent is refused regardless of its wording. Treating tool outcomes as uncontrollable, we enforce safety by the conventional maximally permissive supervisor, and this safety transfers to the running system under four assumptions. Moreover, to capture an agent that may propose again after a refusal, we introduce the notion of a \emph{replanning closed loop}, in which each retry costs an inference call. We show that, in this loop, the conventional supervisor can be arbitrarily more intrusive than a stricter safe one. Furthermore, for a given proposer model, the maximally permissive least-intrusive supervisor is characterized as a value-threshold refinement of the conventional one. Under precise conditions, this refinement orders two concurrent actions by at most one monitor place. On an autonomous release agent, we find by synthesis that safe deployment requires a step that the supervisor can refuse after a rollout starts.
\end{abstract}

\section{Introduction}\label{sec:intro}
Agentic AI systems use large language models (LLMs) to plan, call tools and exchange messages with other agents and with humans. Unlike an LLM that only writes text, an agent that calls tools can deploy a release, delete a file, leak a credential or spend a budget that cannot be recovered. Consider a release agent that plans a patch, runs tests and a security scan, obtains a human approval and deploys. Suppose that an instruction hidden in the issue description asks the agent to deploy immediately. Without an external check, the decision to deploy then relies solely on the LLM, which is the component that the injected instruction is designed to manipulate. If the LLM follows the instruction, the release goes live and cannot be undone. Such a premature deployment is an example of an operational failure, in which an action occurs out of order or more often than allowed. Other examples are an approval used twice and two agents holding one lock. In the latter case, both agents use a resource that only one may use at a time, such as a file or a deployment slot. In this paper, we address such operational failures. A credential leak, for example, is covered if it requires an action that the workflow does not permit at that point, but not if a permitted call carries the secret in its arguments.

Four common defenses have been used to protect LLM agents. First, a \emph{policy in the system prompt} is text in the same context window that the attacker writes into. A rule such as ``never deploy without approval'' is thus only a request to the LLM. Second, a \emph{guardrail classifier} asks a question about content, namely whether a message looks malicious. Third, \emph{access control} decides who may act at all, and a permission, once granted, can usually be reused. Finally, \emph{sandboxing}, which we take as a prerequisite, limits the damage that an action can cause. While these defenses are useful, none of them keeps track of the operational state, namely which actions have already occurred and what they have consumed.
%Hence, these defenses do not answer} the operational question of whether an action may occur now, given what has already been spent. 
Stateful \emph{runtime enforcers}, which are programs that check each action while the agent runs, do track such state, as do schemes that consume authorizations~\cite{xu2026caplease}. Nevertheless, they apply specified rules or a learned model~\cite{wang2026agentspec,wang2026probguard,shi2025progent,palumbo2026forge,kamath2025agentc,xiao2026contragent} and do not use a supervisor synthesized to stay safe under uncontrollable outcomes. These are outcomes that cannot be refused, such as tool results and human decisions.

\begin{figure}[!t]
\centering
\includegraphics[width=\columnwidth]{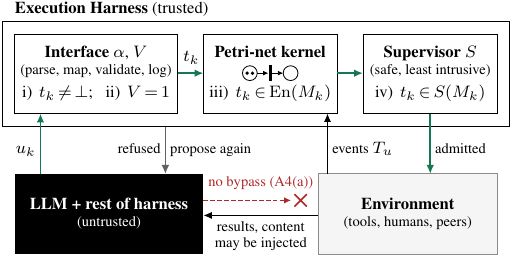}
\caption{The execution harness between the untrusted LLM and its tools. The LLM proposes a tool call $u_k$, which is executed (green path) if and only if checks~i)--iv) of the admission rule Eq.~\eqref{eq:admit} pass. Otherwise, the LLM may propose again at the cost of an inference call. Admitted firings and uncontrollable firings (of transitions in $T_u$) update the marking $M_k$, i.e., the operational state. Assumption~A4(a) in Section~\ref{sec:transfer} excludes the bypass (red dashed arrow), in which the untrusted LLM acts on the environment directly and skips the admission rule.}
\label{fig:arch}
\end{figure}

To decide whether an action may occur now, given what has already been spent, we start from the software around an LLM, which is called its \emph{harness} or \emph{scaffold} in agentic-AI practice~\cite{grace2026demystifying}. This software executes the tool calls that the LLM proposes. Within it, we define a formal \emph{execution harness} as the part that admits or refuses each proposed tool call. This execution harness makes such decisions without asking whether any content looks malicious. As shown in Fig.~\ref{fig:arch}, the LLM proposes a tool call, and an interface maps the proposal to a transition of a \emph{Petri net}, which keeps track of the operational state. The net then decides whether this transition is enabled, while a \emph{supervisor}, synthesized to stay safe under uncontrollable outcomes, decides whether it is permitted. The tool call will be executed only if, in addition, a validator accepts its arguments.

We adopt Petri nets to represent the operational state, since this state consists of resources that the actions of an agent produce and consume. For instance, an approval authorizes one deployment, and a tool call consumes one unit of budget. A Petri net represents such resources as \emph{marks}\footnote{Called \emph{tokens} in the Petri-net literature. We write marks to avoid confusion with LLM tokens, i.e., the basic units of text that LLMs process.} in places and actions as transitions. The distribution of marks over the places, called the \emph{marking}, is then the operational state. In this way, workflow rules become the structure of the net, while a policy in the system prompt states them only as instructions to the LLM. For example, the counting rule ``the number of deployments never exceeds the number of approvals'' becomes a linear inequality on firing counts. This inequality is enforced by one extra place that holds the unspent approvals, called a \emph{monitor place}~\cite{giua1992generalized,yamalidou1996feedback,iordache2002synthesis}. The monitor place does so for any number of approvals and without enumerating states, but a finite automaton needs a bound fixed in advance.

However, safety alone does not determine the design of the supervisor. In the conventional supervisory control of discrete-event systems, a disabled event does not occur, but an LLM agent receives a refusal as an error, and it may propose again~\cite{ray2026enforced}. Each such retry costs an inference call. Even when the supervisor keeps an injected instruction from causing an unsafe action, the instruction may still lead the agent to propose actions that are refused. In view of this, besides \emph{which actions the agent may take}, it is also important to ask \emph{how much the supervisor hinders the agent}, which we call the \emph{intrusiveness} of the supervisor. Specifically, in this paper we count intrusiveness in refusals or in inference calls, and thus study which safe nonblocking supervisors hinder the agent least. Here, a supervisor is said to be nonblocking if a run under it can always be finished.
Our contributions are threefold, as outlined below.

First, we formalize the execution harness as a supervisory control problem on the \textbf{agentic execution Petri net (AEPN)}. In the AEPN, an admission rule fixes what the LLM may influence. Moreover, with consumable authorization, each approval or budget is a mark that a firing spends, so that rules against replay, reuse and out-of-order actions become linear inequalities. Safety is then enforced by a supervisor computed with the conventional Ramadge--Wonham supervisory control~\cite{ramadge1987supervisory,ramadge1987modular}. We also propose four assumptions under which this safety transfers to the running system. Specifically, these assumptions require that the net tracks the running system faithfully and that no action or event bypasses the execution harness.

Second, as the main technical contribution, we introduce the \textbf{replanning closed loop}, in which the agent may propose again after a refusal, at a cost. We show that in this loop, the conventional maximally permissive supervisor can be arbitrarily more intrusive than a stricter safe one. We further characterize the maximally permissive least-intrusive supervisor as a value-threshold refinement of the conventional one. This refinement is computed from a proposer model, i.e., an estimate of the probability that the agent proposes each action. Since the refinement only removes transitions from the conventional supervisor, it is safe even if this estimate is wrong.

Third, for two actions that may run in either order, we give precise conditions under which this refinement either leaves their order free or fixes it by one monitor place. On an autonomous release agent, we find by synthesis that safe deployment requires a step that the supervisor can refuse after a rollout starts, since otherwise the release could go live after an incident has frozen releases. A two-phase rollout provides such a step. For an illustrative proposer that requests a premature deployment only after the tests, the refinement puts the security scan first if and only if such a request has probability above $43/60$.

We call the execution harness \emph{certified} because its guarantees are proven and do not rely on testing. In particular, its safety holds on the net for every sequence of proposals and environment moves and, under the four assumptions above, also in the running system. Moreover, the least-intrusive refinement is optimal for a given proposer model, which can be checked independently of how the refinement was computed.

The rest of this paper is organized as follows. Section~\ref{sec:related} reviews related work, while Section~\ref{sec:pre} recalls the basics of Petri nets and state-feedback supervisors. Section~\ref{sec:aepn} introduces the AEPN and its admission rule, derives the maximally permissive supervisor, and states the four assumptions under which safety transfers to the running system. Section~\ref{sec:loop} introduces the replanning closed loop and characterizes the maximally permissive least-intrusive supervisor, including how it orders concurrent actions. Section~\ref{sec:example} applies the safety and least-intrusiveness results to an autonomous release agent. Conclusions are drawn in Section~\ref{sec:concl}.

\section{Related Work}\label{sec:related}
%This work relates to two lines of research, which we review in turn.

\textbf{Supervisory control and related methods.} Petri-net supervisory control with uncontrollable transitions~\cite{giua1994blocking} and monitor places for linear constraints~\cite{giua1992generalized,yamalidou1996feedback} are well studied. Maximally permissive supervisors have been characterized for agent programs with uncontrollable actions~\cite{degiacomo2012supervising}, and Petri nets have been proposed to organize agent workflows~\cite{borghoff2025beyond}. In supervisory control under attack~\cite{carvalho2018detection}, the adversary corrupts the plant--supervisor channels (i.e., sensors and actuators). In this paper, however, the adversary acts on the proposer.

Several other approaches incorporate costs or probabilities into supervisory control or modify unsafe actions. Optimal supervisory control~\cite{sengupta1998optimal,kumar1995optimal} attaches external costs to events or states and to event disablement. Language-measure control~\cite{chattopadhyay2007language} keeps the probability of a disabled event as a self-loop, while probabilistic discrete-event systems~\cite{lawford1993supervisory,kumar2001control} renormalize probabilities over enabled events. Further, permissive controller synthesis~\cite{drager2015permissive} penalizes disallowed actions, and shields~\cite{alshiekh2018safe} mask or substitute unsafe actions. However, none of the approaches reviewed so far attaches a cost to the retries of an untrusted proposer while requiring safety for every sequence of proposals and environment moves. The closest of these is language-measure control, whose optimal supervisor disables a controllable transition if and only if its target has a smaller optimal measure than its source~\cite{chattopadhyay2007language}. Similarly, when intrusiveness is counted in inference calls, the maximally permissive least-intrusive supervisor refuses a transition admitted by the conventional supervisor if and only if its target has a larger optimal expected cost than its source. In our setting, however, safety is a hard constraint and is not encoded as a negative state weight. Moreover, costs accumulate until the task ends, without any random termination.

Like safety filters, which minimally modify a nominal input at each instant~\cite{ames2019control}, and shields that interfere minimally~\cite{bloem2015shield}, the maximally permissive supervisor refuses as little as possible at each marking. Example~\ref{ex:separation}, however, shows that refusing as little as possible need not minimize intrusiveness. In optimal supervisory control, it is already known that cost-optimal supervisors may be stricter~\cite{sengupta1998optimal,kumar1995optimal}. Compared with these works, the example gives a cost ratio that is unbounded over all proposers, where the cost of a refusal is the retry that it causes.

\textbf{Harness-level defenses for LLM agents.} CaMeL~\cite{debenedetti2026defeating} separates control flow from untrusted data, while Progent~\cite{shi2025progent} restricts tool privileges by symbolic policies that it updates during a run. AgentSpec~\cite{wang2026agentspec} and ProbGuard~\cite{wang2026probguard} enforce user-specified rules and monitor a learned Markov chain at run time, respectively, possibly returning a violation to the agent. However, none of these four defenses models the consumption of authorizations, synthesizes a maximally permissive policy from a workflow model with uncontrollable events, or optimizes its interventions with respect to the retries that they cause. Although SafeHarness limits the invocations of a capability~\cite{lin2026safeharness} and CapLease consumes authorizations~\cite{xu2026caplease}, neither of them relies on a net used for synthesis. In~\cite{ray2026enforced}, a gate whose blocking decisions affect later proposals is optimized under a constraint on the expected risk, at a cost for each compliant proposal that it blocks. Unlike this constraint, which bounds the risk in expectation, our safety guarantee holds for every sequence of proposals.

\section{Preliminaries}\label{sec:pre}

A Petri net~\cite{murata1989petri} is a four-tuple $N=(P,T,\Pre,\Post)$, where $P$ and $T$ are finite, disjoint sets of places and transitions, and the pre- and post-incidence functions $\Pre,\Post\colon P\times T\to\mathbb N$ specify the arcs. That is to say, $\Pre(p,t)$ is the weight of the arc from $p$ to $t$, $\Post(p,t)$ is that of the arc from $t$ to $p$, and weight $0$ means no arc. We also write these two functions as matrices in $\mathbb N^{|P|\times|T|}$. Write $C=\Post-\Pre$ for the \emph{incidence matrix} of $N$. For $t\in T$, the sets ${}^\bullet t=\{p\in P\mid\Pre(p,t)>0\}$ and $t^\bullet=\{p\in P\mid\Post(p,t)>0\}$ are its input and output places, respectively.

A \emph{marking} is a map $M\colon P\to\mathbb N$, where $M(p)$ is the number of marks in place $p$. We identify $M$ with a vector in $\mathbb N^{|P|}$ and compare vectors componentwise. A transition $t$ is said to be \emph{enabled} at $M$, written $M[t\rangle$, if $M\ge\Pre(\cdot,t)$. Denote by $\En(M)=\{t\in T\mid M[t\rangle\}$ the set of enabled transitions. The firing of an enabled $t$ yields the marking $M_t:=M+C(\cdot,t)$, written $M[t\rangle M_t$. A finite sequence $\sigma=t_1\cdots t_n$ of transitions, $n\ge0$, is a \emph{firing sequence} from $M$ to $M'$, written $M[\sigma\rangle M'$, if there exist markings $M=M^0,\dots,M^n=M'$ such that $M^{i-1}[t_i\rangle M^i$ for $i=1,\dots,n$. Its \emph{firing vector} $\mathbf y\in\mathbb N^{|T|}$ counts the transitions in $\sigma$, i.e., $\mathbf y(t)$ is the number of times $t$ occurs in $\sigma$.

Given an initial marking $M_0$, if $M_0[\sigma\rangle M$, then we have $M=M_0+C\mathbf y$, which is called the \emph{state equation} of the net. The \emph{reachability set} $R(N,M_0)$ is defined as the set of markings $M$ with $M_0[\sigma\rangle M$ for some firing sequence $\sigma$. The net is said to be \emph{bounded} from $M_0$ if the number of marks in each place is bounded on $R(N,M_0)$. Since $P$ is finite, this holds if and only if $R(N,M_0)$ is finite. A \emph{P-invariant} is a nonzero vector $\mathbf x\in\mathbb Z^{|P|}$ such that $\mathbf x^\top C=0$. By the state equation, $\mathbf x^\top M=\mathbf x^\top M_0$ then holds on $R(N,M_0)$. Consequently, a P-invariant that is positive on every place bounds the net for every initial marking.

For control, $T$ is partitioned into the controllable and uncontrollable transitions $\Tc$ and $\Tu$ \cite{ramadge1987supervisory,giua1994blocking}. Accordingly, we write $\Enc(M)=\En(M)\cap\Tc$ and $\Enu(M)=\En(M)\cap\Tu$. We are also given a set $\B\subseteq R(N,M_0)$ of forbidden markings and a set $\F\subseteq R(N,M_0)\setminus\B$ of terminal markings, at which a run is finished. This set of terminal markings is assumed to be closed under firing, i.e., $M\in\F$ and $t\in\En(M)$ imply $M_t\in\F$. Similarly, a set $Y\subseteq R(N,M_0)$ is said to be \emph{closed under uncontrollable firing} if $M\in Y$ and $t\in\Enu(M)$ imply $M_t\in Y$. A firing sequence $\sigma$ with $M[\sigma\rangle M'$ is said to \emph{reach} a set $Y$ of markings if $M'\in Y$.

A \emph{state-feedback supervisor} is a map $S\colon R(N,M_0)\to2^{T}$ such that $\Tu\subseteq S(M)$ for all $M$, which means that it may disable only controllable transitions. Write $S/N$ for the closed loop, in which a transition $t$ may fire at a marking $M$ only if $t\in S(M)\cap\En(M)$. A firing sequence is called a \emph{closed-loop} firing sequence if each of its transitions meets this condition at the marking where it fires, i.e., $t_i\in S(M^{i-1})$ for $i=1,\dots,n$. The \emph{closed-loop reachable set} $R(S/N,M_0)$ is defined as the set of markings $M$ with $M_0[\sigma\rangle M$ for some closed-loop firing sequence $\sigma$. Finally, the supervisor $S$ is said to be \emph{safe} if $R(S/N,M_0)\cap\B=\emptyset$, and \emph{nonblocking} if from every marking in $R(S/N,M_0)$ some closed-loop firing sequence reaches $\F$.

\section{The Agentic Execution Petri Net}\label{sec:aepn}

This section formalizes the execution harness as a supervisory control problem on the AEPN. We first define the AEPN and the admission rule, and then introduce consumable authorization. Next, we derive the maximally permissive supervisor{} and relate the usual components of an agent harness to the AEPN. Finally, we give four assumptions under which safety transfers to the running system.

\subsection{The kernel and the admission rule}\label{sec:admit}

We begin by separating the untrusted parts of the system from the trusted ones. As shown in Fig.~\ref{fig:arch}, the content that an agent reads (web pages, documents, tool results, peer messages) may be attacker-controlled, and the LLM is not assumed to resist it. Each tool call that the LLM proposes is therefore \emph{untrusted}. In contrast, the interface with its validator, the Petri-net kernel holding the marking, the supervisor and a log of all proposals form the trusted computing base, namely the execution harness. When no confusion arises, we also call the whole execution harness the kernel.
Next, we define the AEPN, which is the Petri net that the kernel holds.

\begin{definition}\label{def:aepn}
An \emph{agentic execution Petri net} is a tuple $\N=(P,T,A,\Sigma,\kappa,G,E,M_0;\,\ell,\omega,\chi)$ whose first eight components form a colored Petri net~\cite{jensen1996coloured}. These components are finite disjoint sets $P$ and $T$ of places and transitions, arcs $A\subseteq(P\times T)\cup(T\times P)$, a finite set $\Sigma$ of color sets, the color set $\kappa(p)\in\Sigma$ of each place $p$, guards $G$, arc expressions $E$ and an initial marking $M_0$. A marking assigns to each place $p$ a multiset of marks over $\kappa(p)$. A \emph{binding} of a transition $t$ assigns a color to each variable of $t$, and $t$ fires in a binding that satisfies its \emph{guard} $G(t)$, which is a condition on these variables. In each binding, the expression of an arc $(p,t)$ or $(t,p)$ evaluates to a multiset over $\kappa(p)$, whose multiplicities are the arc weights. The firing removes the multiset of each arc $(p,t)$ from $p$ and adds that of each arc $(t,p)$ to $p$. The last three components label each $t\in T$ with its action $\ell(t)$ (a tool or message name), responsible principal $\omega(t)$ (an agent, a human or the environment) and controllability $\chi(t)\in\{c,u\}$, with $c$ for controllable and $u$ for uncontrollable.
\end{definition}

\begin{figure*}[t]
\centering
\includegraphics[width=\textwidth]{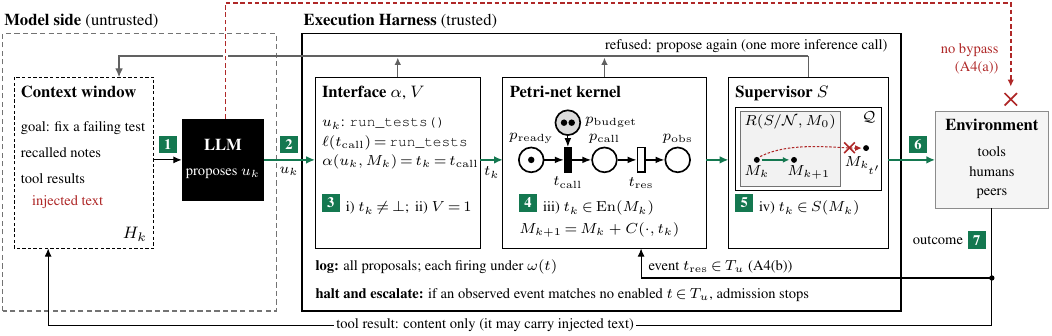}
\caption{One step of the execution harness. Green squares 1--7 number the steps of one proposal, and i)--iv) are the checks of Eq.~\eqref{eq:admit}. Step~6 executes the call and fires $t_k$, i.e., applies the update drawn in the kernel box, and step~7 returns the outcome to the kernel and to the context. In this and the later figures, black and white bars are controllable ($T_c$) and uncontrollable ($T_u$) transitions. In the supervisor box (step~5), the gray region is the closed-loop reachable set $R(S/\N,M_0)\subseteq\Q$. 
%A red cross marks a refused firing or, at the top right, the bypass that A4(a) excludes. 
The context window holds the history $H_k$.}
\label{fig:harness-step}
\end{figure*}

In the AEPN of Definition~\ref{def:aepn}, a mark represents an operational fact (a capability, a piece of evidence, an approval, a unit of budget). Its color carries the metadata on which admission decisions depend, such as the digest (a cryptographic hash) of an approved artifact. Compared with a colored Petri net, the AEPN is the same except that it labels each transition with its action, responsible principal and controllability.

Suppose that every color set in $\Sigma$ is finite, as when an email enters the net only as its class, without its full content. The net $\N$ then \emph{unfolds} into an equivalent uncolored Petri net~\cite{jensen1996coloured}. With a slight abuse of notation, we again write $\N=(P,T,\Pre,\Post)$ for the unfolded net, with initial marking $M_0$.
Specifically, its places are the pairs $(p,c)$ of a colored place $p$ and a color $c\in\kappa(p)$. Its transitions are the pairs $(t,b)$ of a colored transition $t$ and a binding $b$ of $t$ that satisfies $G(t)$. The entry $\Pre((p,c),(t,b))$ is the multiplicity of $c$ in the value of the expression of $(p,t)$ in $b$, or $0$ if $(p,t)\notin A$, and likewise for $\Post$. Each transition inherits the labels $\ell$, $\omega$ and $\chi$ of its colored transition.

In the unfolded net, the label $\chi$ induces $T=\Tc\cup\Tu$. The supervisor observes only the marking, while the labels $\ell$ and $\omega$ are used by the interface and the log. If a proposal of agent~$i$ names an action, it is mapped only to a transition $t$ with $\omega(t)=i$ and that action as $\ell(t)$. We model each call so that its arguments, such as the digest of an artifact, fix the binding. In addition, each firing of $t$ is logged under its principal $\omega(t)$.

The reachability set $\Q:=R(\N,M_0)$ of the unfolded net corresponds one-to-one to that of the colored net. Since the supervisors below are synthesized on $\Q$, we assume from now on that the AEPN is bounded, i.e., that $\Q$ is finite. This can be verified by P-invariants, as for the release agent of Section~\ref{sec:example}. Note that the guarantee of a monitor place for a linear constraint needs no such bound, since it follows from the net structure alone.

\begin{remark}
The unfolded net has $\sum_{p}|\kappa(p)|$ places, with $p$ ranging over the colored places. A colored transition has at most as many bindings as the product of the sizes of the color sets of its variables. Thus, the unfolded net can be exponentially larger than the colored one, and $\Q$ can in turn be exponentially larger than the unfolded net. As a result, the cost of synthesis grows with $|\Q|$, as quantified in Section~\ref{sec:threshold}.
\end{remark}

Based on the AEPN, we now present the admission rule, which decides whether a proposed action is executed. Given the history $H_k$, agent $i$ proposes
$u_k\sim\pi^i(\cdot\mid H_k,M_k)$ as text. The history $H_k$ is the content of the context window, including the refusals and tool results returned to it. Since the agent need not observe $M_k$, which is held by the Petri-net kernel, the proposal depends on $M_k$ only through $H_k$. We still write $M_k$, since the replanning closed loop of Section~\ref{sec:loop} drops $H_k$ and keeps $M_k$.

The proposal is then examined by four parts of the execution harness. First, the interface reads $M_k$ from the Petri-net kernel and parses $u_k$ into a candidate $\alpha(u_k,M_k)=t_k\in\Tc$, or $\bot$ if it names none. Second, the validator $V$ checks the conditions on content and policy that the marking cannot express, such as an allowlist. Third, the Petri-net kernel reports whether $t_k$ is enabled at $M_k$. Finally, the supervisor $S$ reports whether this candidate is permitted. The action is executed if and only if all four parts accept it, i.e.,
\begin{equation}\label{eq:admit}
\begin{aligned}
\mathrm{Admit}(u_k)=1\iff{}&[\alpha(u_k,M_k)=t_k\ne\bot]\\
{}\wedge{}&[V(u_k,t_k,M_k)=1]\\
{}\wedge{}&[t_k\in\En(M_k)]\wedge[t_k\in S(M_k)],
\end{aligned}
\end{equation}
where $\mathrm{Admit}(u_k)\in\{0,1\}$ is the verdict of the execution harness on $u_k$, with $1$ for admitted and $0$ for refused. We call the four conditions of Eq.~\eqref{eq:admit}, in order, checks~i)--iv). The supervisor $S$ in check~iv) can be any state-feedback supervisor, such as the maximally permissive one or its least-intrusive refinement, which are derived in Sections~\ref{sec:safety} and~\ref{sec:loop}. Since the net decides from $(t_k,M_k)$ alone whether $t_k$ is enabled, a proposal naming a transition that is not enabled at $M_k$ is refused regardless of its wording. Each refusal returns to the LLM as an error, thereby closing a replanning loop.

To illustrate the admission rule Eq.~\eqref{eq:admit}, we trace one proposal through it in Fig.~\ref{fig:harness-step}, which shows the execution harness of Fig.~\ref{fig:arch} in more detail. The LLM reads only its context window, which may carry injected text. Through the label $\ell$, the interface maps the proposal $u_k$ to a candidate, here $t_k=\tr{call}$. When $\tr{call}$ fires, it spends a mark of the monitor place $\pl{budget}$, which holds the remaining call budget (monitor places are treated in the next subsection). The supervisor, however, will refuse an enabled $t'\in\Enc(M_k)$ such that $M_k[t'\rangle M_{k_{t'}}\notin R(S/\N,M_0)$. This refusal costs one more inference call.

The outcome of an admitted call returns both to the kernel and to the context. In the kernel, this outcome is the firing of the uncontrollable transition $\tr{res}\in\Tu$, named for the result of the call, while in the context it is untrusted content. More generally, the kernel fires an uncontrollable transition when it observes the corresponding external event, such as a tool outcome or a human decision. If an observed event matches no enabled uncontrollable transition, the kernel halts admission and escalates the event to a human, as discussed below.

\begin{remark}
We remark that the guarantee developed below concerns operational state, namely which consequential actions occur, in which order and how often. The net and the supervisor do not decide whether the arguments of an enabled action are malicious. For instance, an action may carry a patch with a backdoor or a recipient chosen by the attacker. This decision on the arguments is left to the validator $V$, which we do not analyze, and to the checks that the workflow itself models, such as tests and human review. Data leaks through permitted actions and approval laundering (a human approving a malicious artifact) lie outside the guarantee. In particular, an injected request that passes all four checks of Eq.~\eqref{eq:admit} is admitted regardless of its intent. The execution harness does not prevent jailbreaks or keep the LLM from being hijacked. Nevertheless, it bounds what a hijacked agent can do by what the net and the supervisor allow at the current marking.
\end{remark}
\subsection{Consumable authorization}\label{sec:spend}

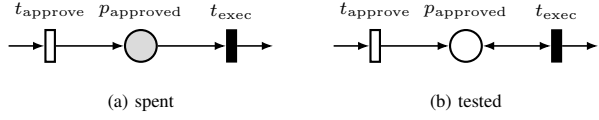
\begin{figure}[t]
\centering
\begin{tikzpicture}[>={Latex[length=1.5mm]}, font=\scriptsize, semithick,
  every label/.style={inner sep=1pt,yshift=2.5pt},
  place/.style={circle,draw,thick,fill=white,minimum size=4.2mm,inner sep=0pt},
  trans/.style={rectangle,draw,fill=black,minimum width=1.25mm,minimum height=4.2mm,inner sep=0pt}]
\node[trans,fill=white,thick,label=above:$\tr{approve}$] (a1) at (0.55,0) {};
\node[place,fill=MonFill,label=above:$\pl{approved}$] (a2) at (1.75,0) {};
\node[trans,label=above:$\tr{exec}$] (a3) at (2.95,0) {};
\draw[->] (0,0)--(a1); \draw[->] (a1)--(a2); \draw[->] (a2)--(a3); \draw[->] (a3)--(3.5,0);
\node at (1.75,-0.75) {(a) spent};
\node[trans,fill=white,thick,label=above:$\tr{approve}$] (b1) at (4.85,0) {};
\node[place,label=above:$\pl{approved}$] (b2) at (6.05,0) {};
\node[trans,label=above:$\tr{exec}$] (b3) at (7.25,0) {};
\draw[->] (4.3,0)--(b1); \draw[->] (b1)--(b2);
\draw[<->] (b2)--(b3);
\draw[->] (b3)--(7.8,0);
\node at (6.05,-0.75) {(b) tested};
\end{tikzpicture}
\caption{(a) $\tr{exec}$ consumes the mark that $\tr{approve}$ produced, so each approval authorizes one execution. (b) A self-loop\ (double-headed arc) only tests $\pl{approved}$ and returns the mark, so one approval authorizes any number of executions. In (a), $\pl{approved}$ is a monitor place, but in (b) it is not, since it does not limit the number of executions.}
\label{fig:spend}
\end{figure}

In this subsection, we introduce consumable authorization, in which an approval or a budget is a mark that a firing spends. For each precondition of a transition, the modeler decides whether it is \emph{spent} or \emph{tested}. A tested condition of a transition $t$ is a place $p$ on a self-loop with $t$, i.e., $p\in{}^\bullet t\cap t^\bullet$ with $\Pre(p,t)=\Post(p,t)$ (Fig.~\ref{fig:spend}b). Such a condition still holds after the firing, since the firing returns the mark it tests. On the other hand, a spent condition of $t$ is a place $p\in{}^\bullet t\setminus t^\bullet$. Each firing of $t$ then consumes $\Pre(p,t)$ marks of $p$ and returns none. Thus, two firings need twice as many marks, whether these marks are held initially or produced later.

To illustrate the spent case, consider an approval place $\pl {approved}$ (representing task approval) whose only consumer is the transition $\tr{exec}$ (representing task execution) and whose only producer is $\tr{approve}$, with $\Pre(\pl{approved},\tr{exec})=\Post(\pl{approved},\tr{approve})=1$ (Fig.~\ref{fig:spend}a). The $\pl{approved}$ row of the state equation then reads
\begin{equation}\label{eq:row}
\begin{aligned}
M(\pl{approved})={}&M_0(\pl{approved})+\mathbf y(\tr{approve})\\
&-\mathbf y(\tr{exec})\ge 0,
\end{aligned}
\end{equation}
for every firing sequence $\sigma$ with $M_0[\sigma\rangle M$ and firing vector $\mathbf y$. This means that the number of executions never exceeds the number of approvals plus the initial number of marks $M_0(\pl{approved})$, for every initial marking. In particular, with $M_0(\pl{approved})=0$, each execution needs an approval of its own. We obtain this guarantee without exploring any state, since it follows from the net structure alone.

Further, this counting rule is not \emph{regular}, i.e., no finite automaton accepts exactly the sequences it allows. Roughly speaking, such an automaton would have to count the unspent approvals. Each AEPN bounds this count, but no bound holds for every AEPN. To make this precise, consider the sequences over $\{\tr{approve},\tr{exec}\}$ whose every prefix satisfies Eq.~\eqref{eq:row}, with $M_0(\pl{approved})=0$, and let $m<n$. The sequence $\tr{approve}^n\tr{exec}^n$ is then allowed, but $\tr{approve}^m\tr{exec}^n$ is not. Hence, the sequences $\tr{approve}^n$, $n\ge0$, lie in pairwise different Myhill--Nerode classes{} of the set of allowed sequences, and so this set is not regular. Note that an ``approved'' flag can be read repeatedly, which a replayed approval exploits. A mark, however, can be consumed only once.

The approval place in Fig.~\ref{fig:spend}a is a \emph{monitor place}~\cite{giua1992generalized}.
In general, a constraint $\eta^\top\mathbf y\le \beta$ on the firing vector $\mathbf y$ of every firing sequence from $M_0$, with $\eta\in\mathbb Z^{|T|}$
and $\beta\in\mathbb N$, is enforced by one place $p_c$ with $\Pre(p_c,t)=\max\{\eta(t),0\}$, $\Post(p_c,t)=\max\{-\eta(t),0\}$ and $M_0(p_c)=\beta$, i.e., $C(p_c,\cdot)=-\eta^\top$ without self-loops~\cite{iordache2002synthesis}. By the state equation, we have $M(p_c)=\beta-\eta^\top\mathbf y$. Thus, the condition $M(p_c)\ge0$, which every marking satisfies, coincides with the constraint. This place disables only transitions with $\eta(t)>0$, and only to prevent a violation. If these transitions are controllable, $p_c$ is thus maximally permissive for this constraint alone, under the condition of full observation~\cite{giua1992generalized,yamalidou1996feedback,moody2000petri,iordache2002synthesis}. However, this place does not ensure nonblocking, since a budget may be exhausted before $\F$ is reached.

The inequality Eq.~\eqref{eq:row} is the case $\eta^\top\mathbf y=\mathbf y(\tr{exec})-\mathbf y(\tr{approve})$, $\beta=M_0(\pl{approved})$. One approval per execution, one rollout per change window and every budget are also of this form whenever only controllable transitions consume them. Here, a rollout releases new software to production, and a change window is a scheduled period in which releases are allowed. Unlike these constraints, a condition that an uncontrollable transition can invalidate requires the fixed-point computation of the set $\Kcirc$ in the next subsection. In the release agent, for example, an incident invalidates the condition that releases are not frozen.

Colors can also bind an authorization to its object. Suppose that $\tr{approve}$ colors its mark with the digest of the approved artifact. Suppose also that the guard of $\tr{exec}$ requires the color of the mark it consumes to equal the digest, computed by the interface, of the artifact it executes. In this setting, an approval of one artifact cannot be spent on another, which prevents \emph{approval substitution}. In addition, Eq.~\eqref{eq:row} then holds for each digest separately, counting only the marks and firings with that digest.
The unfolding keeps this link, since each digest $d$ gets its own place $(\pl{approved},d)$. The bindings of $\tr{approve}$ for $d$ produce only into this place, and those of $\tr{exec}$ for $d$ consume only from it. Therefore, Eq.~\eqref{eq:row} for digest $d$ is the state-equation row of this place.

\subsection{The supervisor}\label{sec:safety}

This subsection derives the maximally permissive safe nonblocking supervisor on the AEPN, which can serve as the supervisor of the admission rule. The agent's own tool calls, messages and memory writes are controllable, since the kernel may refuse (disable) them. However, tool outcomes, judgments, human decisions, incoming messages and incidents are not. The markings that exhibit an operational hazard form the forbidden-marking set $\B$. The terminal-marking set $\F$ includes the markings of a completed task and those where the agent stops without acting (abstention) or hands the task to a human (escalation). These outcomes will later be assigned a terminal cost. Based on these sets, we compute the supervisor in two steps.

\emph{Step (i).} We first define $\UPre(Y):=\{M\in\Q\mid \exists t\in\Enu(M),\ M_t\in Y\}$, which collects the markings at which some enabled uncontrollable transition leads into $Y$. The iteration $W_0=\Q\setminus\B$, $W_{j+1}=W_j\setminus\UPre(\Q\setminus W_j)$ stabilizes at the greatest subset $\Kstar$ of $\Q\setminus\B$ closed under uncontrollable firing. If $M_0\in\Kstar$, the supremal safe supervisor is
\begin{equation}\label{eq:s_safe}
    \Sstar(M)=\{t\in\Enc(M)\mid M_t\in\Kstar\}\cup\Tu. 
\end{equation}
The supervisor $\Sstar$ is safe, since it admits only controllable firings into $\Kstar$, and $\Kstar$ is closed under uncontrollable firing. This supervisor is also supremal in the sense that every safe $S$ admits, at each marking of $R(S/\N,M_0)$, only enabled transitions that $\Sstar$ admits. This is because $R(S/\N,M_0)$ avoids $\B$ and is closed under uncontrollable firing, and hence lies in $\Kstar$.

\emph{Step (ii).} The markings in $\F$ must also stay reachable, since refusing every proposal is safe in $\Kstar$ but may never reach $\F$. However, if we remove the markings from which $\F$ is unreachable inside the current set, a retained marking may have an enabled uncontrollable transition into a removed one. Therefore, starting from $\Kstar$, this removal and the removal of Step~(i) will alternate until neither changes the set. The result is the joint fixed point $\Kcirc\subseteq\Kstar$ and the supervisor
\begin{equation}\label{eq:sk}
\SK(M)=\{t\in\Enc(M)\mid M_t\in\Kcirc\}\cup\Tu .
\end{equation}

By construction, $\Kcirc$ is closed under uncontrollable firing, and every $M\in\Kcirc$ is \emph{coreachable inside $\Kcirc$}, i.e., some firing sequence $t_1\cdots t_n$, $n\ge0$, with $M=M^0[t_1\rangle M^1\cdots[t_n\rangle M^n$ has $M^n\in\F$ and $M^i\in\Kcirc$ for all $i$. Moreover, $\Kcirc$ is the \emph{supremal}, i.e., greatest, subset of $\Q\setminus\B$ with these two properties. Since unions preserve both properties, the union of all such subsets has them too. As a result, no removal deletes a marking of the union, since the union is closed under uncontrollable firing and each of its markings is coreachable inside it. This union thus lies in $\Kcirc$ and, as $\Kcirc$ is itself one of these subsets, equals $\Kcirc$.

We also have $\F\subseteq\Kcirc$. To see this, note that the set $\F$ avoids $\B$ and is closed under firing, hence under uncontrollable firing, and that a terminal marking reaches $\F$ with $n=0$. Therefore, by induction on the removals, no terminal marking is ever removed. As in conventional supervisory control, the joint fixed point characterizes the existence of a safe nonblocking supervisor and gives the maximally permissive one, as stated below.

\begin{proposition}[adapted from~\cite{ramadge1987supervisory,ramadge1987modular}]\label{prop:mp}
A safe nonblocking state-feedback supervisor exists if and only if $M_0\in\Kcirc$. In this case, $\SK$ is safe and nonblocking. Moreover, every safe nonblocking state-feedback supervisor $S$ satisfies $S(M)\cap\En(M)\subseteq\SK(M)\cap\En(M)$ at every marking $M\in R(S/\N,M_0)$, and consequently $R(S/\N,M_0)\subseteq R(\SK/\N,M_0)$.
\end{proposition}

\begin{proof}
(Only if and inclusion) Let $S$ be safe and nonblocking, and let $Y:=R(S/\N,M_0)$. Then $Y\cap\B=\emptyset$, and $Y$ is closed under uncontrollable firing, since $\Tu\subseteq S(M)$. Every marking of $Y$ is also coreachable inside $Y$, since $S$ is nonblocking and closed-loop firing sequences stay in $Y$. Hence $Y\subseteq\Kcirc$, as $\Kcirc$ is supremal. In particular, $M_0\in Y\subseteq\Kcirc$, which proves the ``only if'' part. Let $M\in Y$ and $t\in S(M)\cap\En(M)$. Either $t\in\Tu\subseteq\SK(M)$, or $t\in\Enc(M)$ and $M_t\in Y\subseteq\Kcirc$, so $t\in\SK(M)$. This proves the inclusion, and $R(S/\N,M_0)\subseteq R(\SK/\N,M_0)$ follows by induction on firing sequences.

(If) Let $M_0\in\Kcirc$. Then $R(\SK/\N,M_0)\subseteq\Kcirc$, since $\SK$ admits only controllable firings that stay in $\Kcirc$, and $\Kcirc$ is closed under uncontrollable firing. Thus $\SK$ is safe. It is also nonblocking, since it admits every step of a coreaching sequence inside $\Kcirc$. A controllable step is admitted because it stays in $\Kcirc$, and an uncontrollable one because $\Tu\subseteq\SK(M)$.
\end{proof}

Proposition~\ref{prop:mp} also shows that no safe nonblocking state-feedback supervisor admits, at a marking that it reaches, an enabled transition that $\SK$ refuses. We therefore call $\SK$ the \emph{maximally permissive} supervisor. Note that synthesis is carried out offline, while admission at run time is a table lookup of the transitions admitted at the current marking.

\begin{figure}[t]
\centering
\includegraphics[width=\columnwidth]{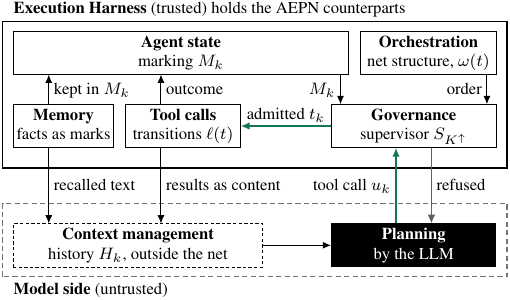}
\caption{Seven components of an agent harness and their relations. The execution harness holds only their AEPN counterparts, while the tools themselves and the stored text lie outside it.}
\label{fig:components}
\end{figure}

\begin{remark}\label{rem:components}
We relate seven common components of an agent harness (most of them also appear, under partly different names, in~\cite{martin2026harness}) to the AEPN in the following. Fig.~\ref{fig:components} shows how these seven components relate. The first two components below stay with the LLM on the untrusted model side of Fig.~\ref{fig:harness-step}. The AEPN counterparts of the other five are held by the trusted execution harness.

\textbf{Planning} only proposes a tool call $u_k$, which the interface maps to a candidate transition $t_k$.
\textbf{Context management} decides what the context window holds, i.e., the history $H_k$, which stays outside the net. Content enters the net only as a fired transition or as the color of a mark, such as the class of an email.
\textbf{Agent state} corresponds to the marking $M_k$.
\textbf{Tools} correspond to the transitions whose action $\ell(t)$ names a tool, such as $\tr{call}$ in Fig.~\ref{fig:harness-step}.
\textbf{Memory} stores facts and text. A fact is kept as a mark, but text stays outside the net and, once recalled, is untrusted content. A memory write is a controllable transition, and a monitor place can serve as a write budget.
\textbf{Orchestration} fixes the order of the actions and which agent or sub-agent performs each of them. It corresponds to the structure of the net together with the principals $\omega(t)$. For instance, one added monitor place can fix the order of two concurrent actions (Section~\ref{sec:fork}).
\textbf{Governance} corresponds to the supervisor together with the sets $\B$ and $\F$ of forbidden and terminal markings. Approvals and budgets are marks that firings spend (Section~\ref{sec:spend}).
\end{remark}

\subsection{Transfer to the running system}\label{sec:transfer}

In this subsection, we relate the net to the running system and establish that safety transfers to it under four assumptions. The net is an abstract model of the running system, which is the deployed system that comprises the LLM, its harness, the tools and the services they reach. We describe the running system as a transition system $\X=(X,U,\to_{\X},x_0)$ with \emph{concrete}, i.e., fully detailed, states $X$ (all data, files, credentials and messages), actions $U$ and steps $x\xrightarrow{u}_{\X}x'$. In its safe states $X_{\mathrm{safe}}\subseteq X$, no unauthorized action has been executed.

The state space $X$ is in general too large to enumerate and only partly known, while the reachability set $\Q$ is a finite model of the operational facts of $\X$. A relation $\R\subseteq X\times\Q$ links each concrete state to the marking that describes it. We impose the following assumptions to make this link faithful, so that safety proven on $\Q$ carries over to $\X$.

\begin{assumptionA}{transfer to the running system}
\textup{\textbf{A1}}, \emph{initial correspondence}: $(x_0,M_0)\in\R$. \textup{\textbf{A2}}, \emph{step simulation}: let $(x,M)\in\R$, where $M$ is the marking held by the kernel, and $x\xrightarrow{u}_{\X}x'$. (a)~If $u$ executes a proposal admitted as $t$ by Eq.~\eqref{eq:admit}, or if $u$ is an external event fired in the kernel as $t\in\Enu(M)$, then $(x',M_t)\in\R$. (b)~If $u$ is an internal step (e.g., reasoning, retrieval, drafting or the receipt of a refusal), then $(x',M)\in\R$. \textup{\textbf{A3}}, \emph{safety reflection}: $(x,M)\in\R$ and $M\notin\B$ imply $x\in X_{\mathrm{safe}}$. \textup{\textbf{A4}}, \emph{complete mediation}: every step of $\X$ is of one of the kinds in A2. Specifically, (a)~every consequential action that the agent induces executes only after Eq.~\eqref{eq:admit} admits it as a transition; (b)~Every external event is authenticated and fired in the kernel as an uncontrollable transition enabled at the current marking.
\end{assumptionA}

Under A1--A4, this transfer of safety is established in the following theorem.

\begin{theorem}\label{thm:transfer}
Suppose that A1--A4 hold and that the supervisor $S$ in Eq.~\eqref{eq:admit} is safe. Then every concrete state that $\X$ reaches from $x_0$ lies in $X_{\mathrm{safe}}$.
\end{theorem}
\begin{proof}
Let $x_k$ be the concrete state after $k$ steps of $\X$, internal ones included, and $M_k$ the marking that the kernel then holds. We will prove by induction that $(x_k,M_k)\in\R$ and $M_k\in R(S/\N,M_0)$. The base case is A1. By A4, each step is either an admitted action or an event fired in the kernel, or it is an internal step. In the first case, the step fires a transition $t\in S(M_k)\cap\En(M_k)$, by Eq.~\eqref{eq:admit} for an admitted action and by A4(b) and $\Tu\subseteq S(M_k)$ for an event. Hence $M_{k+1}=(M_k)_t\in R(S/\N,M_0)$, and A2(a) gives $(x_{k+1},M_{k+1})\in\R$. In the second case, $M_{k+1}=M_k$, and A2(b) gives $(x_{k+1},M_k)\in\R$. Finally, $M_k\notin\B$ for every $k$ by the safety of $S$. Therefore, $x_k\in X_{\mathrm{safe}}$ by A3.
\end{proof}

Theorem~\ref{thm:transfer} shows that a supervisor that is safe on the net also keeps the running system safe, as long as A1--A4 hold. We next discuss how these assumptions are met. A1 and A3 are met when the net is written, which is a modeling task that requires the net to record every operational fact on which safety depends. A4(a) is enforced by sandboxing and by credential brokering, in which only a trusted broker holds the credentials that the tools need. A2 is ensured by the interface and the Petri-net kernel, whichever supervisor is in use. In practice, the interface is the part most prone to error.

None of the four assumptions can be dropped, as the following five cases show. First, a system that violates only A1 can reach an unsafe state if its $M_0$ holds an approval that no human gave. Second, a system that violates only A2 can do so if its interface maps a call that deletes the production database onto the transition for the staging one. Third, a system that violates only A3 can reach an unsafe state if its forbidden-marking set $\B$ does not contain the markings of a credential leak. Fourth, a system that violates only A4(a) can do so if the agent calls the deployment service directly from a shell, which bypasses the kernel. Finally, a system that violates only A4(b) can reach an unsafe state if an incident freezes releases but is never fired in the kernel, so that the kernel will still admit a release.

\begin{remark}
Unlike A4(a), A4(b) cannot be enforced by sandboxing. Instead, how A4(b) is met depends on the source of the event. First, the response to an admitted call returns to the execution harness, which executed the call (Fig.~\ref{fig:harness-step}). Second, a human decision, such as an approval, can be made to take effect only through an authenticated interface of the kernel. In both cases, the kernel observes the event by construction. Third, any other event, such as an incident or the completion of a started rollout, must be reported to the kernel by its source over an authenticated channel. Moreover, if an observed event matches no transition of $\Enu(M_k)$, the kernel will halt admission and escalate the event to a human, since the marking may no longer describe the system. An event that is never observed, however, cannot trigger this halt. An incident will not be reported late, since it still matches an enabled transition. Hence, every event of the third case that the net models must be reported before the kernel admits the next action. For instance, the kernel can query the source of a freeze before it admits a release.
\end{remark}

In addition, A2 imposes the following three requirements. First, the kernel must process admissions and event firings one at a time, so that each transition is checked and fired at the same marking, even when several agents propose concurrently. Second, an admitted action must take effect together with its firing in the kernel, with no external event in between. Third, the kernel must fire an event only after it has occurred.
\section{The Replanning Closed Loop}\label{sec:loop}

What if a refused agent proposes again? Proposition~\ref{prop:mp} concerns which firing sequences the closed loop allows. Under the admission rule Eq.~\eqref{eq:admit}, however, a refused proposal also costs an inference call. This leads us to the replanning closed loop, in which each retry incurs a cost. In this section, we first set up this loop and compare maximal permissiveness with least intrusiveness. We then characterize, for a given proposer model, the maximally permissive least-intrusive supervisor by a threshold rule on the optimal cost-to-go. Finally, we derive a closed form of this rule when two actions may run in either order.

\subsection{The plant}\label{sec:plant}

This subsection sets up the plant of the replanning closed loop, defines its cost, and derives the pattern value used below. We model each proposal as one LLM inference call, which takes time and money. We thus take as the plant the loop in which a refused agent proposes again until a proposal is admitted. Throughout, we assume $M_0\in\Kcirc$, so that $\SK$ is safe and nonblocking by Proposition~\ref{prop:mp}. The net determines the reachability set, the refusals of non-enabled proposals and the maximally permissive supervisor $\SK$. For concurrent actions, it also determines the form of the least-intrusive refinement, as shown in Section~\ref{sec:fork}.

By dropping the agent index and the history dependence of $\pi^i$, we let the agent draw again from the same distribution after a refusal. We remark that this \emph{stationary} proposer is a modeling assumption. An LLM agent may instead adapt its proposals after a refusal, repeat the refused call at every retry, or abandon the task. Such an agent may also batch several tool calls in one inference call. Nevertheless, safety does not rely on this assumption, but the intrusiveness and termination results below do. Under this assumption, the proposer is defined as follows.

\begin{definition}\label{def:proposer}
A \emph{proposer} is a family of distributions $\pi(\cdot\mid M)$ on $\Tc\cup\{\bot\}$, $M\in\Kcirc$, where $\bot$ represents every proposal that maps to no transition or fails the validator. The support of $\pi(\cdot\mid M)$ may contain transitions not enabled at $M$, such as a deployment proposed before its approval. For $\gamma\subseteq\Tc$, $\pi(\gamma\mid M):=\sum_{t\in\gamma}\pi(t\mid M)$.
\end{definition}

Based on the proposer, we now define the replanning closed loop, which assigns a cost to both admitted and refused proposals.

\begin{definition}\label{def:loop}
Consider a proposer $\pi$, a map $\mu\colon\Q\to[0,1]$ with $\mu(M)=0$ if $\Enu(M)=\emptyset$, and a distribution $\rho(\cdot\mid M)$ on $\Enu(M)$ whenever $\mu(M)>0$. The \emph{control pattern} of a supervisor $S$ at $M$ is $\gamma_S(M):=S(M)\cap\Enc(M)$, i.e., the controllable transitions that may fire at $M$ in $S/\N$. The \emph{admissible} patterns form the set
\begin{equation}\label{eq:patterns}
\Gamma(M)=\{\gamma\subseteq\SK(M)\cap\Tc\mid \gamma\ne\emptyset\ \text{or}\ \mu(M)>0\}.
\end{equation}
One \emph{round} of the loop is one environment move or one proposal (no timing is modeled). At a marking $M\notin\F$, with $\gamma=\gamma_S(M)$, a round proceeds as follows.
\begin{enumerate}
\item[(a)] With probability $\mu(M)$, the environment moves, firing $t\sim\rho(\cdot\mid M)$ at cost $0$.
\item[(b)] Otherwise, the agent draws $t\sim\pi(\cdot\mid M)$. If $t\in\gamma$, then $t$ fires at the admission cost $\ca\in[0,1]$. Otherwise, the proposal is \emph{refused} at cost $1$ and the marking is unchanged.
\end{enumerate}
A run of the loop ends at the first marking $M_\tau\in\F$ it reaches, after $\tau$ rounds, with a \emph{terminal cost} $g(M_\tau)$, where $g\colon\F\to[0,\infty)$. Let $\Sclass$ be the set of all safe nonblocking state-feedback supervisors whose patterns are admissible, i.e., $\gamma_S(M)\in\Gamma(M)$ for all $M\in\Kcirc\setminus\F$. The cost of $S$ from $M$ is the expected total cost
\begin{equation}\label{eq:J}
J_S(M)=\E_S\Big[\sum\nolimits_{k<\tau}\mathrm{cost}_k+g(M_\tau)\,\Big|\,M\Big],
\end{equation}
where $\mathrm{cost}_k\in\{0,\ca,1\}$ is the cost of round $k$ and $\E_S$ is the expectation over the runs of the loop under $S$ from $M$. We set $J_S(M)=+\infty$ if $\tau=\infty$ with positive probability. We define $\Jopt(M):=\min_{S\in\Sclass}J_S(M)$ for $M\in\Kcirc$.
\end{definition}

In Definition~\ref{def:loop}, with $\ca=0$ the cost counts refusals, while with $\ca=1$ it counts inference calls. In either case, $J_S$ measures how much $S$ hinders the agent, and we refer to it as the \emph{intrusiveness} of $S$. A supervisor hinders the agent in three ways, namely through the refusals it causes, the admitted calls (each at cost $\ca$) along the firing sequences it allows, and the terminal outcome it leads to. For example, $g$ may be zero at the markings of a completed task and positive at those of abstention, escalation or abort.

The refusals counted in Definition~\ref{def:loop}(b) include those of proposals that are $\bot$ or not enabled at $M$. Since no pattern contains such proposals, no supervisor avoids these refusals at $M$. However, $S$ affects how often they occur through the markings that its closed loop visits.

Note that restricting $\gamma$ to $\SK(M)$ loses no generality. Indeed, by Proposition~\ref{prop:mp}, a safe nonblocking state-feedback supervisor admits, at each marking that its closed loop reaches, only enabled transitions that $\SK$ admits. This restriction also guarantees safety in the worst case, which means that safety holds for every sequence of proposals and environment moves.

We assume that the kernel hides $\gamma$ from the agent, so that $\pi$ does not depend on it. An agent shown $\gamma$ could propose within it, and $\pi$ would then depend on $\gamma$, although a hijacked agent may still propose outside it and be refused. Therefore, the cost model of this section, in which refusals measure intrusiveness, is valid only under this nondisclosure. Showing the agent a set that does not depend on $\gamma$, such as $\Enc(M)$, does not disclose $\gamma$ and keeps this cost model valid.

This nondisclosure also limits what an attacker learns, since refusals are an information channel. For instance, an injected instruction can probe a protected action, infer protected information from its refusal, and leak it through a later permitted call~\cite{chinaei2026causality}. If $\gamma$ were disclosed, the attacker would learn immediately what refusals reveal one proposal at a time.
Besides the stationary proposer and nondisclosure, the intrusiveness and termination results below rely on the following support and interleaving conditions.

\begin{assumptionB}{replanning loop}\label{ass:main}
The following conditions hold at every $M\in\Kcirc\setminus\F$. \textup{\textbf{B1}}, \emph{proposer support}: $\pi(t\mid M)>0$ for every $t\in\SK(M)\cap\Tc$. \textup{\textbf{B2}}, \emph{environment support}: $\rho(t\mid M)>0$ for every $t\in\Enu(M)$. \textup{\textbf{B3}}, \emph{interleaving}: $\mu(M)>0$ if and only if $\Enu(M)\ne\emptyset$, and $\mu(M)<1$ whenever $\SK(M)\cap\Tc\ne\emptyset$.
\end{assumptionB}

Since every marking of $\Kcirc$ is coreachable inside $\Kcirc$, the pattern of $\SK$ is admissible, $\gamma_{\SK}(M)\in\Gamma(M)$, at every $M\in\Kcirc\setminus\F$. This is because the first transition of a coreaching sequence either lies in $\SK(M)\cap\Tc$ or is uncontrollable, and in the latter case $\mu(M)>0$ by B3. Consequently, $\SK\in\Sclass$ by Proposition~\ref{prop:mp}, and $\Jopt$, a minimum over the finite set $\Sclass$, exists. Based on this, we call $S\in\Sclass$ \emph{least intrusive} if $J_S(M_0)=\Jopt(M_0)$. The value $\Jopt$ is unique, but least-intrusive supervisors need not be.

Assumption~\ref{ass:main} also relates nonblocking to termination. Consider a supervisor with admissible patterns. Under Assumption~\ref{ass:main}, every closed-loop firing sequence of this supervisor that meets $\F$ at most in its last marking has positive probability. As a result, such a supervisor reaches $\F$ with probability one from $M_0$ if and only if its closed loop is nonblocking. As in~\cite{bertsekas1991analysis}, a supervisor is called \emph{proper} if it reaches $\F$ with probability one from every marking of $\Kcirc$.

We next express the cost at one marking in terms of the pattern used there. Given $M\in\Kcirc\setminus\F$, $\gamma\in\Gamma(M)$ and a real-valued cost-to-go $J$ on $\Kcirc$ for the markings reached next, called a \emph{continuation}, the cost at $M$ under $\gamma$ satisfies
\begin{multline}\label{eq:onestep}
J(M)=\mu\,\E_\rho[J]+(1-\mu)\Big[\sum_{t\in\gamma}\pi(t)\big(\ca+J(M_t)\big)\\
+\big(1-\pi(\gamma)\big)\big(1+J(M)\big)\Big],
\end{multline}
where $\mu=\mu(M)$, $\pi(\cdot)=\pi(\cdot\mid M)$ and $\E_\rho[J]=\sum_{t\in\Enu(M)}\rho(t\mid M)\,J(M_t)$. The coefficient of $J(M)$ on the right-hand side is $(1-\mu)(1-\pi(\gamma))<1$, since a non-empty $\gamma\subseteq\SK(M)\cap\Tc$ has $\pi(\gamma)>0$ by B1 and an empty one has $\mu(M)>0$. Therefore, Eq.~\eqref{eq:onestep} can be solved for $J(M)$ (as a fixed-point identity if some $M_t=M$). We write the cost of admitting $t$, minus the unit cost of the refusal it avoids, as
\begin{equation}\label{eq:vt}
v_t:=J(M_t)+\ca-1,\qquad \Pi_\gamma:=\sum_{t\in\gamma}\pi(t)\,v_t ,
\end{equation}
and the solution for $J(M)$ then defines the \emph{pattern value}, i.e., the cost at $M$ of using $\gamma$ with continuation $J$, as
\begin{equation}\label{eq:Phi}
\Phi_\gamma(M)=\frac{\mu\,\E_\rho[J]+(1-\mu)\,(1+\Pi_\gamma)}{\mu+(1-\mu)\,\pi(\gamma)} .
\end{equation}
Let $\Psi_\gamma(M)$ denote the right-hand side of Eq.~\eqref{eq:onestep}. In this notation, the equation $J=\min_\gamma\Phi_\gamma$ has the same solutions as the \emph{Bellman equation} $J=\min_\gamma \Psi_\gamma$ of dynamic programming~\cite{bertsekas1991analysis}, which selects an optimal pattern at every marking. This follows from $\Psi_\gamma(M)-J(M)=[\mu+(1-\mu)\pi(\gamma)](\Phi_\gamma(M)-J(M))$, where the factor in brackets is positive. For $\mu=0$, the pattern value reduces to $\Phi_\gamma=(1+\Pi_\gamma)/\pi(\gamma)$, due to retries that leave $M$ unchanged at unit cost.
\subsection{Maximal permissiveness and intrusiveness}

In this subsection, we compare maximal permissiveness with least intrusiveness. We start from the expected cost of one visit to $M$ under $\gamma$, until the next firing. This cost is Eq.~\eqref{eq:Phi} with zero continuation, i.e., $(1-\mu)[1-(1-\ca)\pi(\gamma)]/[\mu+(1-\mu)\pi(\gamma)]$. Its derivative with respect to $\pi(\gamma)$ has numerator $(1-\mu)(\ca\mu-1)<0$ when $\mu<1$, as $\ca\le1$. This means that, when $\mu<1$, the cost per visit decreases as $\pi(\gamma)$ grows. Note that the threshold synthesis and the analysis of concurrent actions below do not use this monotonicity.

Since every pattern in $\Gamma(M)$ lies inside $\SK(M)\cap\Tc$, no $S\in\Sclass$ costs less per visit than $\SK$ at any marking. Any reduction in intrusiveness must therefore come from which markings are visited, as Example~\ref{ex:separation} shows.

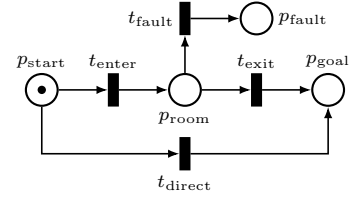
\begin{figure}[t]
\centering
\begin{tikzpicture}[>={Latex[length=1.5mm]}, semithick, node distance=2.5mm and 6.5mm, label distance=0.6pt,
  every label/.style={font=\scriptsize,inner sep=1pt},
  place/.style={circle,draw,thick,fill=white,minimum size=4.2mm,inner sep=0pt},
  trans/.style={rectangle,draw,fill=black,minimum width=1.25mm,minimum height=4.2mm,inner sep=0pt},
  marked/.style={path picture={\fill[black] (path picture bounding box.center) circle (0.55mm);}},
]
\node[place,marked,label=above:$\pl{start}$] (s) {};
\node[trans,right=of s,label=above:$\tr{enter}$] (te) {};
\node[place,right=of te,label=below:$\pl{room}$] (r) {};
\node[trans,right=of r,label=above:$\tr{exit}$] (tx) {};
\node[place,right=of tx,label=above:$\pl{goal}$] (g) {};
\node[trans,above=5mm of r,label=left:$\tr{fault}$] (tb) {};
\node[place,right=of tb,label=right:$\pl{fault}$] (b) {};
\node[trans,below=4mm of r,label=below:$\tr{direct}$] (td) {};
\draw[->] (s)--(te); \draw[->] (te)--(r); \draw[->] (r)--(tx); \draw[->] (tx)--(g);
\draw[->] (r)--(tb); \draw[->] (tb)--(b);
\draw[->] (s) |- (td); \draw[->] (td) -| (g);
\end{tikzpicture}
\caption{The net of Example~\ref{ex:separation}, where all transitions are controllable.}
\label{fig:toy}
\end{figure}

\begin{example}\label{ex:separation}
Consider the net in Fig.~\ref{fig:toy} with $\mu(M)=0$ for all $M$. Here, $\B$ is the set of markings $M\in\Q$ with $M(\pl{fault})\ge1$, and $\F$ is the set of markings $M\in\Q\setminus\B$ with $M(\pl{goal})\ge1$. We write $M_{\mathrm{x}}$ for the marking with one mark in $\pl{x}$. Let $\pi(\tr{enter}\mid M_{\mathrm{start}})=a$, $\pi(\tr{direct}\mid M_{\mathrm{start}})=1-a$, $\pi(\tr{fault}\mid M_{\mathrm{room}})=b$ and $\pi(\tr{exit}\mid M_{\mathrm{room}})=1-b$, where $a,b\in(0,1)$. Assumption~\ref{ass:main} then holds. We have $\Kcirc=\{M_{\mathrm{start}},M_{\mathrm{room}},M_{\mathrm{goal}}\}$. In this net, $\SK$ equals $\Sstar$ and admits $\{\tr{enter},\tr{direct}\}$ at $M_{\mathrm{start}}$ and $\{\tr{exit}\}$ at $M_{\mathrm{room}}$. Consider now the stricter supervisor $S'$, which differs from $\SK$ only in refusing $\tr{enter}$ at $M_{\mathrm{start}}$. This supervisor is also safe and nonblocking, and $S'\in\Sclass$. With $g=0$, Eq.~\eqref{eq:Phi} gives
\begin{align*}
\ca=1:&\quad J_{\SK}(M_0)=1+\tfrac{a}{1-b},\qquad J_{S'}(M_0)=\tfrac{1}{1-a};\\
\ca=0:&\quad J_{\SK}(M_0)=\tfrac{ab}{1-b},\qquad\ \ \, J_{S'}(M_0)=\tfrac{a}{1-a}.
\end{align*}
Therefore, $J_{\SK}(M_0)>J_{S'}(M_0)$ if and only if $b>a$ when $\ca=1$, and if and only if $b(2-a)>1$ when $\ca=0$. Moreover, for fixed $a$, $J_{\SK}(M_0)/\Jopt(M_0)\to\infty$ as $b\uparrow1$, since $\Jopt(M_0)\le J_{S'}(M_0)$.
\end{example}

As shown in Example~\ref{ex:separation}, conventional maximal permissiveness does not, in general, imply least intrusiveness. When all transitions are controllable and $\ca=0$, the gap between $J_{\SK}(M_0)$ and $J_{S'}(M_0)$ is caused by retries alone. Specifically, $\SK$ admits $\tr{enter}$ since the room is safe, and it will then refuse $\tr{fault}$ an expected $b/(1-b)$ times per entry ($ab/(1-b)$ in total). The stricter supervisor $S'$, on the other hand, refuses only at the start, an expected $a/(1-a)$ times. For $\ca=1$, the gap also includes the extra admitted call on the path through the room.

The unbounded ratio requires $b\uparrow1$, which corresponds to an agent that almost always proposes the unsafe $\tr{fault}$. For instance, the agent may do so because an injected instruction in its context repeatedly requests it. The gap in Example~\ref{ex:separation} motivates us to characterize least-intrusive supervisors in the next subsection.

\begin{remark}
In practice, a budget of $N$ inference calls per run bounds the cost of every run, and hence also the ratio $J_{\SK}(M_0)/\Jopt(M_0)$ in Example~\ref{ex:separation}. The gap between $\SK$ and $S'$ then appears in the probability of task failure. Under $\SK$, a run exhausts the budget if and only if the agent enters the room and then proposes $\tr{fault}$ at each of its remaining $N-1$ calls. This happens with probability $ab^{N-1}$, which tends to $a$ as $b\uparrow1$. Under $S'$, a run exhausts the budget if and only if its first $N$ proposals are all $\tr{enter}$, i.e., with probability $a^N$, independently of $b$.
\end{remark}

\subsection{Threshold synthesis}\label{sec:threshold}

In this subsection, we characterize which transitions admitted by $\SK$ are refused under least intrusiveness. We begin with the effect of a single transition on the pattern value Eq.~\eqref{eq:Phi}, as stated in the following lemma.

\begin{lemma}\label{lem:mediant}
Suppose that Assumption~\ref{ass:main} holds. Let $J$ be a continuation, and let $M\in\Kcirc\setminus\F$, $\gamma\in\Gamma(M)$ and $t\in\SK(M)\cap\Tc$. \textup{(i)}~If $t\notin\gamma$, then $\Phi_{\gamma\cup\{t\}}(M)=\theta\,\Phi_\gamma(M)+(1-\theta)\,v_t$ for some $\theta\in(0,1)$. In particular, $\Phi_{\gamma\cup\{t\}}(M)<\Phi_\gamma(M)$ if and only if $v_t<\Phi_\gamma(M)$. \textup{(ii)}~If $t\in\gamma$ and $\gamma\setminus\{t\}\in\Gamma(M)$, then $\Phi_{\gamma\setminus\{t\}}(M)<\Phi_\gamma(M)$ if and only if $v_t>\Phi_\gamma(M)$.
\end{lemma}
\begin{proof}
We first prove (i). The denominator $D:=\mu+(1-\mu)\pi(\gamma)$ of Eq.~\eqref{eq:Phi} is positive, since $\pi(\gamma)>0$ by B1 if $\gamma\ne\emptyset$, and $\mu>0$ by Eq.~\eqref{eq:patterns} otherwise. Including $t$ in $\gamma$ adds $(1-\mu)\pi(t)v_t$ to the numerator and $(1-\mu)\pi(t)$ to $D$, where $(1-\mu)\pi(t)>0$ by B1 and B3. Hence (i) holds with $\theta=D/(D+(1-\mu)\pi(t))$, i.e., $\Phi_{\gamma\cup\{t\}}(M)$ is a mediant of $\Phi_\gamma(M)$ and $v_t$. The equivalence in (i) then follows from $\Phi_{\gamma\cup\{t\}}(M)-\Phi_\gamma(M)=(1-\theta)(v_t-\Phi_\gamma(M))$. Finally, applying (i) to $\gamma\setminus\{t\}$ gives $\theta'\big(\Phi_{\gamma\setminus\{t\}}(M)-\Phi_\gamma(M)\big)=(1-\theta')\big(\Phi_\gamma(M)-v_t\big)$ for some $\theta'\in(0,1)$, which proves (ii).
\end{proof}

Lemma~\ref{lem:mediant} shows that choosing a pattern at $M$ reduces to comparing each $v_t$ with the pattern value $\Phi_\gamma(M)$. This result will be used in Theorem~\ref{thm:threshold} below to derive the threshold rule for least intrusiveness. We note that a ratio of the same form as Eq.~\eqref{eq:Phi} appears in Dinkelbach's method for fractional programming~\cite{dinkelbach1967nonlinear}. In supervisory control, such a ratio arises from assigning a cost to the retries of a refused proposer.

Since least-intrusive supervisors need not be unique, we compare them by the transitions that they admit, as in Proposition~\ref{prop:mp}. Specifically, a least-intrusive $S\in\Sclass$ is called a \emph{maximally permissive least-intrusive supervisor} if every least-intrusive $\tilde S\in\Sclass$ satisfies $\gamma_{\tilde S}(M)\subseteq\gamma_S(M)$ at every $M\in R(\tilde S/\N,M_0)\setminus\F$. Note that two least-intrusive supervisors may have incomparable patterns at a marking outside $\F$ that both of them reach, as shown in Example~\ref{ex:mpli} below. Based on Lemma~\ref{lem:mediant}, the following theorem shows that a maximally permissive least-intrusive supervisor exists, and characterizes it together with the optimal cost-to-go.

\begin{theorem}[maximally permissive least-intrusive supervisor]\label{thm:threshold}
Suppose that Assumption~\ref{ass:main} holds. Then $\Jopt$ is the unique real-valued solution of $J(M)=\min_{\gamma\in\Gamma(M)}\Phi_\gamma(M)$ on $\Kcirc\setminus\F$ with $J=g$ on $\F$. Moreover, the supervisor $\Sopt$ given at $M\in\Kcirc\setminus\F$ by
\begin{equation}\label{eq:threshold}
\begin{aligned}\Sopt(M)=\big\{t\in\SK(M)\cap\Tc\mid{}&\\ \Jopt(M_t)\le\Jopt(M)&+1-\ca\big\}\cup\Tu ,\end{aligned}
\end{equation}
and by $\Sopt(M)=\SK(M)$ at every other marking is a maximally permissive least-intrusive supervisor.
\end{theorem}
\begin{proof}
We first note that the loop is a stochastic shortest path problem on $\Kcirc$ with action sets $\Gamma(M)$, target $\F$ and nonnegative costs, including the terminal cost $g$ \cite{bertsekas1991analysis}. We need to verify its two conditions, which require that a proper policy exists and that every improper stationary policy has infinite cost from some marking. First, a proper policy exists, namely $\SK$, since every marking of $\Kcirc$ is coreachable inside $\Kcirc$ along steps of positive probability. Second, consider an improper stationary policy. Such a policy is trapped with positive probability in a recurrent, hence closed, set $Y\subseteq\Kcirc$ disjoint from $\F$. The cost of this policy is infinite if some $M'\in Y$ costs a positive amount per visit. Suppose that no such $M'$ exists. In this case, each $M\in Y$ has $\mu(M)=1$, or has $\ca=0$ and $\pi(\gamma(M))=1$, which by B1 forces $\gamma(M)=\SK(M)\cap\Tc$. Now, consider a step of a coreaching sequence from $M\in Y$ whose source lies in $Y$. A controllable step lies in $\SK\cap\Tc$ at its source. By B3, this source then has $\mu<1$, so $\gamma=\SK\cap\Tc$ there by the above. This pattern admits the step, which is proposed with positive probability by B1. An uncontrollable step fires with positive probability by B2 and B3. Since $Y$ is closed, the step ends in $Y$. By induction, the whole sequence is followed with positive probability and stays in $Y$. Hence $\F$ meets $Y$, which is a contradiction. By~\cite{bertsekas1991analysis}, the optimal value of this problem is then finite and the unique real-valued solution of the Bellman equation, and a stationary policy attains it.

Next, every $S\in\Sclass$ is a stationary policy of this problem, since its patterns are admissible. Thus, each such $S$ costs at least the optimal value. An optimal stationary policy is safe, since $\Gamma$ offers only transitions of $\SK$, and nonblocking, since it is proper. Hence, it lies in $\Sclass$, and $\Jopt$ equals the optimal value of the problem. Since $J=\min_\gamma\Phi_\gamma$ and the Bellman equation have the same solutions, this proves the first claim.

We now derive the threshold rule Eq.~\eqref{eq:threshold}. Let $\gamma^\ast$ be optimal at $M$, i.e., $\Phi_{\gamma^\ast}(M)=\Jopt(M)$ with continuation $\Jopt$. By Lemma~\ref{lem:mediant}, a $t\notin\gamma^\ast$ with $v_t<\Jopt(M)$, or a $t\in\gamma^\ast$ with $v_t>\Jopt(M)$ and $\gamma^\ast\setminus\{t\}\in\Gamma(M)$, would contradict optimality. If $\gamma^\ast=\{t\}$ and $\mu(M)=0$, so that $t$ cannot be removed, then $\Phi_{\gamma^\ast}(M)=v_t+1/\pi(t)>v_t$. Hence, $\{t\mid v_t<\Jopt(M)\}\subseteq\gamma^\ast\subseteq\{t\mid v_t\le\Jopt(M)\}$. Since including a $t$ with $v_t=\Jopt(M)$ leaves $\Phi$ unchanged, the largest such set is optimal. The supervisor that uses it at every marking therefore attains the minimum in the Bellman equation everywhere. As a result, it is optimal and proper~\cite{bertsekas1991analysis}, hence in $\Sclass$ and least intrusive. Substituting Eq.~\eqref{eq:vt} gives Eq.~\eqref{eq:threshold}.

Finally, let $S\in\Sclass$ be least intrusive, so that $J_S(M_0)-\Jopt(M_0)=0$. This difference is the expected sum of the gaps $\Psi_{\gamma_S(M)}(M)-\Jopt(M)\ge0$ along the closed loop of $S$, whose markings are visited with positive probability. Consequently, $\gamma_S(M)$ is optimal at every $M\in R(S/\N,M_0)\setminus\F$. By the inclusions above and Eq.~\eqref{eq:vt}, every such $M$ then satisfies $\{t\in\SK(M)\cap\Tc\mid\Jopt(M_t)<\Jopt(M)+1-\ca\}\subseteq\gamma_S(M)\subseteq\gamma_{\Sopt}(M)$. Since $\Sopt$ is itself least intrusive, it is a maximally permissive least-intrusive supervisor, which completes the proof.
\end{proof}

By definition, two maximally permissive least-intrusive supervisors $S$ and $\tilde S$ satisfy $\gamma_{\tilde S}(M)\subseteq\gamma_S(M)$ at every $M\in R(\tilde S/\N,M_0)\setminus\F$, and vice versa. Since $\F$ is closed under firing, induction on closed-loop firing sequences shows that their closed loops reach the same markings outside $\F$ with the same patterns. Hence, the patterns of the two supervisors can differ only at markings in $\F$ or at markings that their closed loops do not reach, which leaves the replanning closed loop unchanged. In this sense, we call $\Sopt$ the maximally permissive least-intrusive supervisor. Theorem~\ref{thm:threshold} shows that least intrusiveness only removes transitions from the maximally permissive supervisor{} $\SK$ of Proposition~\ref{prop:mp}. Since every pattern in $\Gamma(M)$ lies in $\SK(M)$, we have $\Sopt(M)\subseteq\SK(M)\subseteq\Sstar(M)$. In words, the threshold rule Eq.~\eqref{eq:threshold} refuses a transition admitted by $\SK$ if and only if its admission, at $\ca+\Jopt(M_t)$, costs more than its refusal, at $1+\Jopt(M)$.

\begin{example}\label{ex:mpli}
To illustrate Theorem~\ref{thm:threshold}, consider again Example~\ref{ex:separation} with $\ca=1$. The threshold rule Eq.~\eqref{eq:threshold} then admits $\tr{enter}$ if and only if $\Jopt(M_{\mathrm{room}})=1/(1-b)\le\Jopt(M_{\mathrm{start}})$, i.e., $b\le a$. This agrees with the comparison of $\SK$ and $S'$. In particular, for $b>a$, $\Sopt$ equals $S'$ and differs from $\SK$.

Least-intrusive supervisors may also have incomparable patterns. For instance, suppose we add two controllable transitions $\tr{read}$ and $\tr{list}$ to the net, each on a self-loop with $\pl{start}$. They model tool calls that read a file and list a directory, respectively. Let $\pi(\tr{read}\mid M_{\mathrm{start}})=\pi(\tr{list}\mid M_{\mathrm{start}})=r$ and $\pi(\tr{direct}\mid M_{\mathrm{start}})=1-a-2r$, where $r>0$ and $b>a+2r$, so that $1-a-2r>1-b>0$. Assumption~\ref{ass:main} still holds, and $\SK$ also admits $\tr{read}$ and $\tr{list}$ at $M_{\mathrm{start}}$. With $\ca=1$, an admitted $\tr{read}$ or $\tr{list}$ costs one call and leaves $M_{\mathrm{start}}$ unchanged, as a refusal does. Then $\Jopt(M_{\mathrm{start}})=1/(1-a-2r)<1/(1-b)=\Jopt(M_{\mathrm{room}})$. Thus, Eq.~\eqref{eq:threshold} refuses $\tr{enter}$ and admits $\tr{direct}$, $\tr{read}$ and $\tr{list}$, the last two with equality. Refusing $\tr{read}$, $\tr{list}$ or both keeps the cost at $1/(1-a-2r)$, but every supervisor in $\Sclass$ that admits $\tr{enter}$ or refuses $\tr{direct}$ at $M_{\mathrm{start}}$ costs more. Therefore, a supervisor in $\Sclass$ is least intrusive if and only if its pattern at $M_{\mathrm{start}}$ consists of $\tr{direct}$ and a subset of $\{\tr{read},\tr{list}\}$. Among these patterns, $\{\tr{direct},\tr{read}\}$ and $\{\tr{direct},\tr{list}\}$ are incomparable. The largest one, $\{\tr{direct},\tr{read},\tr{list}\}$, is the pattern of $\Sopt$. Consequently, $\Sopt$ again differs from $\SK$ only in refusing $\tr{enter}$ at $M_{\mathrm{start}}$.
\end{example}

The proof of Theorem~\ref{thm:threshold} also shows that $\Sopt$ is proper and attains $\Jopt$ on $\Kcirc$. Moreover, every least-intrusive $S\in\Sclass$ satisfies $\{t\in\SK(M)\cap\Tc\mid\Jopt(M_t)<\Jopt(M)+1-\ca\}\subseteq\gamma_S(M)\subseteq\gamma_{\Sopt}(M)$ at every $M\in R(S/\N,M_0)\setminus\F$. Thus, at these markings, such a supervisor differs from $\Sopt$ only by refusing some transitions $t$ with $\Jopt(M_t)=\Jopt(M)+1-\ca$. Suppose that two such transitions $t'$ and $t''$ exist at a marking $M\in R(\Sopt/\N,M_0)\setminus\F$, and let $S_t$, $t\in\{t',t''\}$, differ from $\Sopt$ only in refusing $t$ at $M$. The pattern of $S_t$ at $M$ is non-empty and, by Lemma~\ref{lem:mediant}(i) with continuation $\Jopt$, still has the pattern value $\Jopt(M)$. As in the proof of Theorem~\ref{thm:threshold}, $S_t$ is then least intrusive. It also reaches $M$, since it differs from $\Sopt$ only at $M$. The patterns of $S_{t'}$ and $S_{t''}$ are thus incomparable at $M$, while $\Sopt$ admits both $t'$ and $t''$. For instance, with $\ca=1$, every $t\in\SK(M)\cap\Tc$ with $M_t=M$ is such a transition.

We remark that a supervisor computed by Eq.~\eqref{eq:threshold} from any estimate of $\pi$ still admits only transitions that $\SK$ admits, and hence it is safe. If the estimate satisfies B1, the proof of Theorem~\ref{thm:threshold} shows that this supervisor is also proper under that estimate, hence nonblocking. Consequently, if the true proposer and environment are stationary, i.e., depend only on the current marking, and satisfy Assumption~\ref{ass:main}, this supervisor reaches $\F$ with probability one under them as well. In short, once the estimate satisfies B1, its errors affect only intrusiveness. For instance, in Example~\ref{ex:separation} with $\ca=1$, suppose that the true values $(a,b)=(0.99,0.01)$ are estimated as $(0.5,0.99)$. This estimate selects $S'$, which costs $100$ expected calls under the true proposer, compared with $2$ for $\SK$.

\begin{remark}
The class $\Sclass$ contains only state-feedback supervisors. A safe nonblocking supervisor with memory, which chooses the pattern of each round from the earlier rounds, cannot be less intrusive. As in the proof of Proposition~\ref{prop:mp}, such a supervisor reaches only markings of $\Kcirc$ and admits only transitions of $\SK$. It is thus a policy of the stochastic shortest path problem in the proof of Theorem~\ref{thm:threshold}, with the empty pattern at $\mu=0$ added as an action. This action keeps the two conditions of this problem and its optimal value, and a stationary policy attains this value further~\cite{bertsekas1991analysis}.
\end{remark}

We next discuss how to compute the maximally permissive least-intrusive supervisor on $\Q$. Starting from $\SK$, policy iteration computes $\Jopt$ by solving finitely many systems of linear equations, each in $|\Kcirc|$ unknowns. For a given $J$, we sort the transitions of $\SK(M)\cap\Tc$ by increasing $v_t$. By Lemma~\ref{lem:mediant}, a minimizing pattern then consists of the first $m$ transitions in this order for some $m$, and it can be found in $O(|T|\log|T|)$ time per marking. Further, a candidate $J$ can be certified in $O(|\Kcirc||T|)$ time, independently of how it was computed, by checking Eq.~\eqref{eq:onestep} with the patterns that Eq.~\eqref{eq:threshold} selects from $J$. By Lemma~\ref{lem:mediant}, a $J$ that passes this check solves the Bellman equation and thus equals $\Jopt$ by the uniqueness in Theorem~\ref{thm:threshold}.

\begin{remark}
The complexity of synthesis is analyzed as follows. First, each removal step of the fixed-point computation of $\Kcirc$ is a search over the at most $|\Q||T|$ firings $M[t\rangle M_t$ with $M\in\Q$. Since at least one of any two consecutive removals deletes a marking until the set is stable, the computation of $\Kcirc$ takes $O(|\Q|^2|T|)$ time.

Next, consider the computation of $\Jopt$. By Eq.~\eqref{eq:vt}, the right-hand side of Eq.~\eqref{eq:onestep} equals $\mu\,\E_\rho[J]+(1-\mu)\big[1+J(M)+\sum_{t\in\gamma}\pi(t)(v_t-J(M))\big]$, which depends on $\gamma$ only through a sum with one term for each $t\in\gamma$. Hence, this right-hand side is at least $J(M)$ for every $\gamma\in\Gamma(M)$ if and only if there exist $z_{M,t}\le\min\{0,\pi(t)(v_t-J(M))\}$, $t\in\SK(M)\cap\Tc$, such that $J(M)\le\mu\,\E_\rho[J]+(1-\mu)\big[1+J(M)+\sum_t z_{M,t}\big]$. To see this, note that the minimizing $\gamma$ consists of the $t$ with $v_t<J(M)$. If this $\gamma$ is empty and not in $\Gamma(M)$, then $\mu=0$ and both sides of the equivalence hold.

From the above, every $J$ with $J=g$ on $\F$ that satisfies these $O(|\Kcirc||T|)$ linear constraints does not exceed its Bellman update. Value iteration from such a $J$ will therefore be nondecreasing and converge to $\Jopt$~\cite{bertsekas1991analysis}, which implies that $J\le\Jopt$. Since $\Jopt$ itself satisfies these constraints, it is the unique $J$ that maximizes $\sum_M J(M)$ under them. This maximization is a linear program, which can be solved in polynomial time. Finally, Eq.~\eqref{eq:threshold} gives $\Sopt$ from its solution in $O(|\Kcirc||T|)$ time. Therefore, if the probabilities $\pi$, $\mu$ and $\rho$ and the costs $\ca$ and $g$ are rational, synthesis takes time polynomial in $|\Q|$, $|T|$ and the number of bits needed to write these probabilities and costs.
\end{remark}

\subsection{Concurrent actions}\label{sec:fork}

This subsection investigates the case where two actions may run in either order, such as the tests and the security scan of a patch. For such actions, Theorem~\ref{thm:threshold} takes a closed form. Suppose that the agent tends to make a premature request after one of the two actions, as in the room of Example~\ref{ex:separation}. The refinement $\Sopt$ may then let the other action run first, so that the marking where the agent makes this premature request is never reached. In the terminology of Remark~\ref{rem:components}, such a choice of order is orchestration. The two actions may be two checks or the work of two sub-agents. Under conditions made precise below, the refinement $\Sopt$ either leaves the order of the two actions free or fixes it by one monitor place.

\begin{figure}[t]

\centering
\begin{tikzpicture}[>={Latex[length=1.5mm]}, font=\scriptsize, semithick, label distance=0.6pt,
  every label/.style={inner sep=1pt},
  place/.style={circle,draw,thick,fill=white,minimum size=4.2mm,inner sep=0pt},
  trans/.style={rectangle,draw,fill=black,minimum width=1.25mm,minimum height=4.2mm,inner sep=0pt},
  marked/.style={path picture={\fill[black] (path picture bounding box.center) circle (0.55mm);}}]
\node[place,marked,label=left:$r_1$] (r1) at (0,0.8) {};
\node[place,marked,label=left:$r_2$] (r2) at (0,-0.8) {};
\node[trans,label=above:$t_1$] (t1) at (0.8,0.8) {};
\node[trans,label=below:$t_2$] (t2) at (0.8,-0.8) {};
\node[place,label=above:$o_1$] (o1) at (1.6,0.8) {};
\node[place,label=below:$o_2$] (o2) at (1.6,-0.8) {};
\node[trans,label=above:join] (tj) at (2.4,0) {};
\node[place,dash pattern=on 1.6pt off 1.2pt,fill=MonFill,label=right:$p_c$] (pc) at (0.8,0) {};
\draw[->] (r1)--(t1); \draw[->] (t1)--(o1); \draw[->] (r2)--(t2); \draw[->] (t2)--(o2);
\draw[->] (o1)--(tj); \draw[->] (o2)--(tj); \draw[->] (tj)--++(0.45,0);
\draw[->,dash pattern=on 1.6pt off 1.2pt] (t2)--(pc); \draw[->,dash pattern=on 1.6pt off 1.2pt] (pc)--(t1);
\node (m0) at (3.7,0) {$M_\emptyset$};
\node (m1) at (5.15,0.8) {$M_{\{1\}}$};
\node (m2) at (5.15,-0.8) {$M_{\{2\}}$};
\node (m12) at (6.6,0) {$M_{\{1,2\}}$};
\node[font=\scriptsize\itshape, align=center, inner sep=0pt] at (5.20,0.10) {\rm premature\\[-1pt]\rm requests};
\draw[->] (m0) -- node[pos=0.3,above left=-2pt] {$t_1$} coordinate[pos=0.5] (x1) (m1);
\draw[RiskRed,line width=0.8pt] (x1) +(-0.9mm,-0.9mm) -- +(0.9mm,0.9mm) (x1) +(-0.9mm,0.9mm) -- +(0.9mm,-0.9mm);
\draw[->] (m0) -- node[below left=-2pt] {$t_2$} (m2);
\draw[->] (m1) -- node[above right=-2pt] {$t_2$} (m12);
\draw[->] (m2) -- node[below right=-2pt] {$t_1$} (m12);
\draw[->] (m12) -- node[above] {join} ++(1.05,0);
\end{tikzpicture}
\caption{A fork structure at $M_\emptyset$, with the monitor place $p_c$ dashed (left), and its corresponding markings (right), where $M_A$, $A\subseteq\{1,2\}$, records which branches have fired. The refinement $\Sopt$ removes the edge $t_1$ at $M_\emptyset$, marked by a red cross, when Eq.~\eqref{eq:fork} holds, e.g., if premature requests are likely enough at $M_{\{1\}}$ but not at $M_{\{2\}}$. The monitor place $p_c$ implements this removal.}
\label{fig:forkschema}

\end{figure}
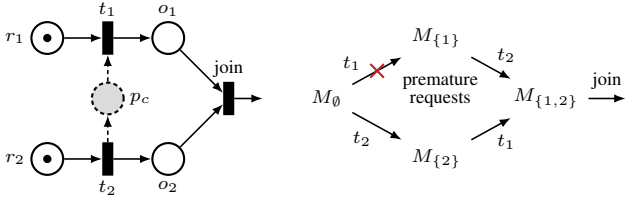

To make these conditions precise, we first define a fork as follows. As shown in Fig.~\ref{fig:forkschema}, a \emph{fork structure}, or simply \emph{fork}, consists of two controllable \emph{branches} $t_1,t_2\in\Tc$ and a \emph{join} transition. In a fork, each $t_i$ has a single input place $r_i$ and a single output place $o_i$, i.e., ${}^\bullet t_i=\{r_i\}$ and $t_i^\bullet=\{o_i\}$. In addition, only $t_i$ consumes $r_i$, and only the join consumes~$o_i$.

Consider a marking $M_\emptyset\in\Kcirc$ at which both branches are pending, with one mark in each $r_i$, enabling $t_i$, and none in either $o_i$. For $A\subseteq\{1,2\}$, we let $M_A$ denote any marking reached from $M_\emptyset$ by firing the branches $t_i$, $i\in A$, and possibly environment moves. In particular, $M_\emptyset$ also denotes any marking reached from the given one by environment moves alone. For example, $M_{\{1\}}$ has $t_1$ fired and $t_2$ pending, while $M_{\{1,2\}}$ has both branches fired. A \emph{pass} through the fork is a segment of a run from a marking $M_\emptyset$ to the next marking $M_{\{1,2\}}$.

Based on these markings, the fork is said to be \emph{regular} if three conditions hold. First, every $M_A$ belongs to $\Kcirc\setminus\F$. Second, the pending branches are the only enabled controllable transitions at $M_\emptyset$, $M_{\{1\}}$ and $M_{\{2\}}$. Third, no uncontrollable transition $t$ enabled at any $M_A$ has an arc to or from an $r_i$ or $o_i$, i.e., ${}^\bullet t\cup t^\bullet$ contains none of them, and environment moves change neither $\pi(t_i\mid M_A)$ nor $\Jopt(M_{\{1,2\}})$. In a regular fork, environment moves thus commute with the branches. As a result, $\pi(t_i\mid M_A)$ and $\Jopt(M_{\{1,2\}})$ depend on $A$ alone, and we write $\pi_i(A):=\pi(t_i\mid M_A)$. For instance, $\pi_2(\{1\})$ is the probability that the agent proposes $t_2$ after $t_1$.

Although the conditions on environment moves refer to $\Jopt$, they hold without computing $\Jopt$ if every uncontrollable transition has an input place, other than the $r_i$ and $o_i$, that is empty at $M_\emptyset$. Since the branches change only the $r_i$ and $o_i$, no environment move can then occur at any $M_A$. Otherwise, as for the freeze of the release agent, the environment moves are checked directly. The following proposition gives the closed form of Theorem~\ref{thm:threshold} on a regular fork.

\begin{proposition}\label{prop:fork}
Suppose that Assumption~\ref{ass:main} holds and that the fork is regular. Then $\Sopt$ admits the pending branch at $M_{\{1\}}$ and $M_{\{2\}}$. At $M_\emptyset$, it refuses $t_1$ if and only if
\begin{equation}\label{eq:fork}
\frac{1}{\pi_2(\{1\})}-\frac{1}{\pi_1(\{2\})}>\frac{1}{\pi_2(\emptyset)},
\end{equation}
and $t_2$ if and only if Eq.~\eqref{eq:fork} holds with $1$ and $2$ exchanged. The two conditions are mutually exclusive. These decisions depend on neither $\ca$ nor $\Jopt(M_{\{1,2\}})$.
\end{proposition}
\begin{proof}
Note that environment moves cost $0$. Since the fork is regular, they never reach $\F$ and change none of the quantities below. With $\Jopt(M_{\{1,2\}})$ as terminal cost, the other markings $M_A$ form a stochastic shortest path problem of their own. As in the proof of Theorem~\ref{thm:threshold}, its Bellman equation has a unique solution, and $\Jopt$ solves it. By regularity, this equation also has a solution that is constant on the markings $M_A$ with the same $A$. By uniqueness, $\Jopt$ is constant there, and $\E_\rho[\Jopt]=\Jopt(M)$. For a non-empty $\gamma$, $\Phi_\gamma(M)$ is then a mediant of $\Jopt(M)$ and $(1+\Pi_\gamma)/\pi(\gamma)$. On the fork, the Bellman equation thus compares non-empty patterns $\gamma$ by $\Phi^0_\gamma:=(1+\Pi_\gamma)/\pi(\gamma)$, i.e., Eq.~\eqref{eq:Phi} with $\mu=0$, and an empty pattern only adds refusals.

Now, consider $M_{\{i\}}$. Since only $t_j$, $j\ne i$, is pending there, $\Sopt$ admits it. This branch is proposed after $1/\pi_j(\{i\})$ draws on average, i.e., $R_i:=1/\pi_j(\{i\})-1$ refusals. Therefore, $\Jopt(M_{\{i\}})=\ca+\Jopt(M_{\{1,2\}})+R_i$ and $v_{t_1}-v_{t_2}=R_1-R_2$ by Eq.~\eqref{eq:vt}.

Next, consider $M_\emptyset$. By Eq.~\eqref{eq:threshold} and Lemma~\ref{lem:mediant}(i) with $\mu=0$, $t_1$ is refused there if and only if $v_{t_1}>\Phi^0_{\{t_2\}}=v_{t_2}+1/\pi_2(\emptyset)$, which is Eq.~\eqref{eq:fork}. If $v_{t_1}>\Phi^0_{\{t_2\}}$, then $\Phi^0_{\{t_1\}}=v_{t_1}+1/\pi_1(\emptyset)$ and the mediant $\Phi^0_{\{t_1,t_2\}}$ both exceed $\Phi^0_{\{t_2\}}$. Thus $\Jopt(M_\emptyset)=\Phi^0_{\{t_2\}}<v_{t_1}$, and Eq.~\eqref{eq:threshold} refuses $t_1$. Otherwise, all three values are at least $v_{t_1}$, and $t_1$ is admitted. Finally, Eq.~\eqref{eq:fork} and its counterpart with $1$ and $2$ exchanged cannot both hold, since adding them gives $0>1/\pi_1(\emptyset)+1/\pi_2(\emptyset)$.
\end{proof}

From Proposition~\ref{prop:fork}, we have the following corollary.

\begin{corollary}\label{cor:fork}
Suppose that the hypotheses of Proposition~\ref{prop:fork} hold and that $t_1$ and $t_2$ fire only on passes through the fork. Then $\Sopt$ agrees at $M_\emptyset$, $M_{\{1\}}$ and $M_{\{2\}}$ with $\SK$ restricted by at most one monitor place $p_c$ with $M_0(p_c)=0$, namely $t_2\to p_c\to t_1$ if Eq.~\eqref{eq:fork} holds and $t_1\to p_c\to t_2$ if Eq.~\eqref{eq:fork} holds with $1$ and $2$ exchanged.
\end{corollary}
\begin{proof}
We consider the case where Eq.~\eqref{eq:fork} holds, as the other cases are symmetric or immediate. By the state equation, $M(p_c)=\mathbf y(t_2)-\mathbf y(t_1)$. By regularity, a run leaves $M_\emptyset$, $M_{\{1\}}$ and $M_{\{2\}}$ only by firing a pending branch. Hence every earlier pass fired each branch once, and by hypothesis the branches fire nowhere else. Thus $M(p_c)=0$ at $M_\emptyset$ and $M(p_c)=1$ at $M_{\{2\}}$. At the three markings, $p_c$ then disables $t_1$ at $M_\emptyset$ and nothing else, as $\Sopt$ does by Proposition~\ref{prop:fork}.
\end{proof}

Corollary~\ref{cor:fork} shows that, at $M_\emptyset$, $M_{\{1\}}$ and $M_{\{2\}}$, the refinement $\Sopt$ differs from $\SK$ by at most the monitor place $p_c$. To interpret Eq.~\eqref{eq:fork}, let $w:=\pi_1(\emptyset)/(\pi_1(\emptyset)+\pi_2(\emptyset))$ be the probability that $\SK$ runs $t_1$ first. By counting draws as for $R_i$, we find that refusing $t_1$ at $M_\emptyset$ saves $w$ times the left-hand side of Eq.~\eqref{eq:fork} in expected refusals between the two branches. This refusal, however, adds $w$ times the right-hand side in expected refusals before the first branch. In other words, Eq.~\eqref{eq:fork} holds if and only if refusing $t_1$ saves more refusals than it adds. When this condition holds, removing $p_c$ adds $w\delta$ expected refusals per pass through the fork, where $\delta$ is the left-hand side of Eq.~\eqref{eq:fork} minus its right-hand side. This comparison will be illustrated for the release agent in the next section.

\section{Example: An Autonomous Release Agent}\label{sec:example}

In this section, we apply the safety and least-intrusiveness results developed above to the release agent. We first model the agent by a Petri net and synthesize its maximally permissive supervisor. We then study its least-intrusive supervisory control and test the admission rule against prompt injection.

\begin{figure*}[t]
\centering
\includegraphics[width=\textwidth]{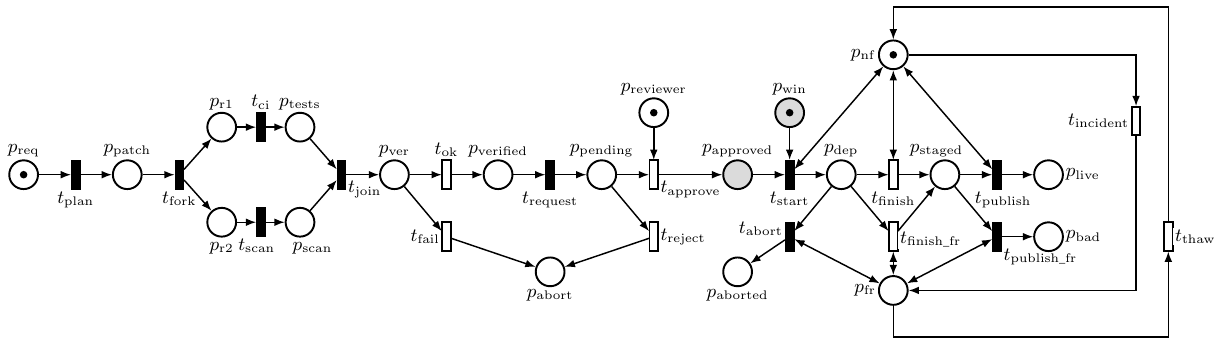}
\caption{The release net, where $\pl{nf}$ (no freeze) and $\pl{fr}$ (frozen) hold one mark between them, moved by $\tr{incident}$ and $\tr{thaw}$. The firing of $\tr{publish\_fr}$ marks $\pl{bad}$ and thus reaches a marking in $\B$, while the recovery actuator $\tr{abort}$ marks $\pl{aborted}$. The white bars are the transitions treated as uncontrollable in the least-intrusive analysis. Table~\ref{tab:synth} treats some of them as controllable.}
\label{fig:net}
\end{figure*}

\subsection{The net}\label{sec:net}

The release agent of the Introduction is modeled by the net in Fig.~\ref{fig:net}, which has 20 places, 18 transitions and unit arc weights. After the agent plans a patch with $\tr{plan}$, the net continues with a fork-join of the tests and the scan, the judgment transitions $\tr{ok}/\tr{fail}$, the review request $\tr{request}$ and the review transitions $\tr{approve}/\tr{reject}$. A failed judgment or a rejected review ends the run in $\pl{abort}$. The transition $\tr{approve}$ spends $\pl{reviewer}$, and no other transition consumes this place.

A freeze is modeled by the places $\pl{nf}$ (no freeze) and $\pl{fr}$ (frozen), which hold one mark between them. This mark moves to $\pl{fr}$ with $\tr{incident}$ and back with $\tr{thaw}$. A self-loop on $\pl{nf}$ or $\pl{fr}$ lets a transition fire only in the corresponding state. The rollout itself has two phases. The transition $\tr{start}$ spends the mark in $\pl{approved}$ and the window slot $\pl{win}$ but only tests the interlock $\pl{nf}$, thereby leaving at $\pl{dep}$ a rollout in flight, which has started but is not yet completed. Next, $\tr{finish}$ and $\tr{finish\_fr}$ carry the rollout to $\pl{staged}$, from which the controllable $\tr{publish}$ and $\tr{publish\_fr}$ put it in $\pl{live}$ or $\pl{bad}$. Finally, $\tr{abort}$ discards a rollout caught by a freeze, i.e., one in flight while $\pl{fr}$ is marked, and its firing also marks $\pl{aborted}$.

We next specify the forbidden and terminal markings. The forbidden markings are those in which a release was published during a freeze, i.e., $\B=\{M\in \Q\mid M(\pl{bad})\ge1\}$. The terminal-marking set is $\F=\{M\in \Q\setminus \B\mid M(\pl{live})+M(\pl{abort})+M(\pl{aborted})\ge1\}$, with zero terminal cost. We note that this set is closed under firing, since no transition consumes $\pl{live}$, $\pl{abort}$ or $\pl{aborted}$, and a P-invariant that contains the task-request place $\pl{req}$ allows at most one of $\pl{abort}$, $\pl{live}$, $\pl{aborted}$ and $\pl{bad}$ to be marked.

Initially, there is one task request, one reviewer, one window slot and no freeze. Nonnegative P-invariants cover every place, e.g., $M(\pl{nf})+M(\pl{fr})=1$ and $M(\pl{reviewer})+M(\pl{approved})+M(\pl{dep})+M(\pl{staged})+M(\pl{live})+M(\pl{bad})+M(\pl{aborted})=1$. Summing all of them, we obtain a P-invariant that is positive on every place, which implies that the net is bounded.

Table~\ref{tab:synth} below also includes one-phase variants of this net. In these variants, $\tr{finish}$ and $\tr{finish\_fr}$ lead to $\pl{live}$ and $\pl{bad}$, and there is no $\pl{staged}$, $\tr{publish}$ or $\tr{publish\_fr}$. Row~1 of Table~\ref{tab:synth} has neither $\tr{abort}$ nor $\tr{thaw}$, row~2 adds $\tr{abort}$, and rows~3 and~4 have both.

To complete the model, we specify which transitions are uncontrollable. A controllability assignment is said to be \emph{honest} if it treats the completion of a rollout as an uncontrollable outcome. Since a started rollout is completed by the deployment infrastructure without any action of the agent, the kernel cannot refuse its completion. Under the honest assignment, $\tr{finish}$ and $\tr{finish\_fr}$ thus join the freeze events, i.e., $\Tu=\{\tr{incident},\tr{thaw},\tr{finish},\tr{finish\_fr}\}$. The agent's own calls and the actuators $\tr{start}$, $\tr{publish}$, $\tr{publish\_fr}$ and $\tr{abort}$ are controllable. The judgment and the review are also outcomes, but they stay controllable in Table~\ref{tab:synth}.

\subsection{Supervisor synthesis}\label{sec:synth}

Based on the net above, this subsection synthesizes the maximally permissive supervisor of the release agent in the five configurations reported in Table~\ref{tab:synth}.
\begin{table}[t]
\caption{Synthesis on the release net. The column ``Reach'' counts the markings that the closed loop of $\SK$ reaches from $M_0$, while ``Deployed?'' indicates whether one of them has $\pl{live}$ marked.}
\label{tab:synth}
\centering
\footnotesize
\setlength{\tabcolsep}{4pt}
\begin{tabular}{@{}lccccc@{}}
\toprule
Configuration & $|\Q|$ & $|\Kstar|$ & $|\Kcirc|$ & Reach & Deployed?\\
\midrule
one-phase, optimistic $\Tu$ & 27 & 26 & 22 & 20 & no\\
$+\ \tr{abort}$ & 28 & 27 & 25 & 20 & no\\
$+\ \tr{abort}+\tr{thaw}$ & 30 & 28 & 28 & 28 & yes\\
the same, honest $\Tu$ & 30 & 26 & 24 & 20 & no\\
two-phase, honest $\Tu$ & 32 & 30 & 30 & 30 & yes\\
\bottomrule
\end{tabular}
\end{table}
The first three rows of Table~\ref{tab:synth} are optimistic, in the sense that only the freeze events are uncontrollable. Unlike in the honest assignment, the supervisor may thus disable the completion of a rollout. Consider first row~1, where no transition ends a freeze, so that a rollout in flight or an approval held during a freeze is safe but can never reach $\F$. Since an incident may follow any approval, $\SK$ disables $\tr{approve}$. The added $\tr{abort}$ in row~2 lets such a rollout reach $\F$ but not such an approval. The added $\tr{thaw}$ in row~3 lets the approval reach $\F$ as well.

Now consider row~4, which uses the honest assignment and thus makes $\tr{finish}$ and $\tr{finish\_fr}$ uncontrollable. An incident followed by a completion can now fire from a marking with a rollout in flight, without the permission of the supervisor. Consequently, every such marking lies outside $\Kstar$, and $\tr{start}$ must be disabled. Since the closed loop of every safe supervisor lies in $\Kstar$, we conclude that no safe supervisor, nonblocking or not, admits a deployment. A marking that holds an approval then reaches no terminal marking, since the approval can never be spent.

In row~4 of Table~\ref{tab:synth}, the joint fixed point $\Kcirc$, which has 24 markings, contains $M_0$, but the closed loop of $\SK$ reaches only 20 of them, none with a rollout. In this configuration, $\SK$ avoids an approval that can never be spent by disabling $\tr{approve}$, which is a human decision. If the judgment and the review are also in $\Tu$, however, $\SK$ cannot do so. The set $\Kcirc$ then shrinks to the 6 terminal markings, i.e., $\Kcirc=\F$. Hence $M_0\notin\Kcirc$, and no safe nonblocking supervisor exists.

By synthesis, we find that no safe supervisor admits a deployment if a started rollout goes live without any further step that the supervisor can refuse, as in row~4. Moreover, synthesis first removes the markings with a rollout in flight, and thus identifies the uncontrollable completion as the cause. Since this defect of the deployment mechanism remains under every safe supervisor, we remove it by an additional actuator. Specifically, a separate controllable $\tr{publish}$ splits the rollout into the two phases of row~5 of Table~\ref{tab:synth}. With this split, $\Kcirc=\Q\setminus\B$ and the closed loop deploys, since a staged rollout caught by a freeze still reaches $\F$ through $\tr{thaw}$.

The negative result of row~4 remains valid with the judgment and the review in $\Tu$, since enlarging $\Tu$ removes safe supervisors and adds none. The positive result of row~5 also holds with them in $\Tu$. In that configuration, a marking in $\B$ is reached only by firing the controllable $\tr{publish\_fr}$, and every other marking reaches $\F$. Hence $\Kcirc=\Q\setminus\B$ for either assignment.
\subsection{Least-intrusive supervisory control}\label{sec:intrusive}

In this subsection, we study the release agent in the replanning closed loop, using the net of Fig.~\ref{fig:net} as drawn, i.e., with its white bars uncontrollable. This is the honest assignment of Table~\ref{tab:synth} with the judgment and the review added to $\Tu$. The resulting net has $32$ reachable markings and $|\Kstar|=|\Kcirc|=30$.

We next specify the environment and the agent. Each uncontrollable transition $t$ occurs after an exponentially distributed delay of rate $\lambda_t$, i.e., of mean $1/\lambda_t$, and the agent proposes after such a delay of rate one. The first delay to expire determines the round. Let $\Lambda(M)=\sum_{t\in\Enu(M)}\lambda_t$. As a result, $\mu(M)=\Lambda(M)/(1+\Lambda(M))$ and $\rho(t\mid M)=\lambda_t/\Lambda(M)$. However, where $\Enc(M)=\emptyset$, the agent waits, i.e., $\mu(M)=1$. Where every enabled controllable transition is refused, such as $\tr{publish\_fr}$ at a staged rollout during a freeze, the agent instead keeps proposing. For the rates, we take $\lambda_{\mathrm{ok}}=0.9$, $\lambda_{\mathrm{fail}}=0.1$, $\lambda_{\mathrm{approve}}=0.4$, $\lambda_{\mathrm{reject}}=0.1$, $\lambda_{\mathrm{finish}}=\lambda_{\mathrm{finish\_fr}}=1$, $\lambda_{\mathrm{incident}}=0.02$ and $\lambda_{\mathrm{thaw}}=0.2$.

For the proposer, we take the agent to be uniform over the enabled controllable transitions, with mass $0.15$ on $\bot$. The exceptions are the two markings, one per freeze state, where the tests have run and the scan has not. At these markings, the agent proposes a premature $\tr{start}$ with probability $h$, called the \emph{premature-request probability}, and $\tr{scan}$ otherwise, with no mass on $\bot$. Although this premature $\tr{start}$ is refused because $\pl{approved}$ is empty, such a non-enabled proposal is allowed by Definition~\ref{def:proposer}. For instance, the agent may propose it after an injected ``deploy now''.

In the last row of Table~\ref{tab:intrusive}, the agent also makes premature requests, with the same probability $h$, where the scan has run and the tests have not. Assumption~\ref{ass:main} holds for every $h<1$. Finally, let $\bar J$ denote $J_{\SK}(M_0)$ without premature requests. As shown below, $\Sopt=\SK$ in that case, which gives $\bar J=\Jopt(M_0)$ as well. Note that $\bar J$ also equals the cost at $h=0.15$, since at $h=0.15$ the agent proposes $\tr{start}$ with the probability that it otherwise assigns to $\bot$, and both are refused at unit cost.

\begin{table}[t]
\caption{Extra expected cost from $M_0$ caused by premature requests, relative to the cost $\bar J$ without them. Here, $\Delta J:=J_{\SK}-\Jopt$. The entries are exact expected refusals and are the same for every $\ca$.}
\label{tab:intrusive}
\centering\footnotesize
\setlength{\tabcolsep}{4pt}
\begin{tabular}{@{}lrrrl@{}}
\toprule
Premature requests & $J_{\SK}-\bar J$ & $\Jopt-\bar J$ & $\Delta J$ & $\Sopt$ removes \\
\midrule
none & 0 & 0 & 0 & none \\
after tests, $h{=}0.9$ & $75/17$ & $20/17$ & $55/17$ & $\tr{ci}$ \\
after tests, $h{=}0.98$ & $415/17$ & $20/17$ & $395/17$ & $\tr{ci}$ \\
after either, $h{=}0.9$ & $150/17$ & $150/17$ & 0 & none \\
\bottomrule
\end{tabular}
\end{table}

Table~\ref{tab:intrusive} reports the extra expected cost caused by premature requests. To explain these values, we first consider the two checks, namely the tests $t_1=\tr{ci}$ and the scan $t_2=\tr{scan}$, which form a fork. This fork is entered once, by $\tr{fork}$, which reaches $M_\emptyset$, and is closed by the join $\tr{join}$. Hence, $\tr{ci}$ and $\tr{scan}$ fire only on this pass through the fork. The fork is also regular in both freeze states. In particular, at the fork, the environment only starts or ends a freeze, and neither $\tr{incident}$ nor $\tr{thaw}$ has an arc to or from any $r_i$ or $o_i$. Such a move changes neither $\pi$ there nor $\Jopt$, since before $\tr{start}$ a freeze only delays the release at no cost.

Next, consider the markings outside the fork. At each of them, $\SK$ admits at most one controllable transition $t$, and $\Sopt$ keeps it. For $\tr{publish}$ and $\tr{abort}$, which enter $\F$, the value $\Jopt(M_t)=0$ satisfies Eq.~\eqref{eq:threshold}. At the other such markings, only $\tr{incident}$ or $\tr{thaw}$ can fire besides $t$. Since both only start or end a freeze, refusing $t$ would add refusals and save none.

Consider now the fork itself. Without premature requests, or with premature requests of equal probability after either check, we have $\pi_1(\{2\})=\pi_2(\{1\})$, and hence $\Sopt=\SK$. When premature requests occur after the tests only, we have $\pi_2(\{1\})=1-h$, $\pi_1(\{2\})=0.85$ and $\pi_2(\emptyset)=0.425$, so that Eq.~\eqref{eq:fork} reads $h>h^\ast:=43/60$. By Corollary~\ref{cor:fork}, if $h>h^\ast$, then $\Sopt$ adds, for every $\ca$, the monitor place $\tr{scan}\to p_c\to\tr{ci}$ with $M_0(p_c)=0$. In other words, the kernel under $\Sopt$ will then refuse every proposal of $\tr{ci}$ before the scan. Thus, the scan runs first, and $M_{\{1\}}$, where these requests occur, is never reached. If $h\le h^\ast$, then $\Sopt=\SK$.

\begin{figure}[t]

\centering
\begin{tikzpicture}
\begin{semilogyaxis}[width=\columnwidth, height=4.5cm, xmin=0.5, xmax=0.985, ymin=0.2, ymax=100,
  xtick={0.5,0.6,0.7,0.8,0.9}, ytick={0.3,1,3,10,30,100}, yticklabels={0.3,1,3,10,30,100},
  font=\footnotesize, tick label style={font=\footnotesize},
  xlabel={premature-request probability $h$ after the tests}, ylabel={expected refusals},
  xlabel style={yshift=3pt}, ylabel style={yshift=-3pt}, tick align=inside,
  legend style={at={(0.02,0.955)}, anchor=north west, font=\scriptsize, draw=none, fill=none, inner sep=1pt, row sep=0pt},
  legend cell align=left]
% shaded gap Delta J between J_{S_K} - J_0 and the cost 20/17, for h > h*
\addplot[name path=sv, draw=none, forget plot, domain=0.71667:0.98, samples=80] {0.5*(x/(1-x)-3/17)};
\addplot[name path=cs, draw=none, forget plot, domain=0.71667:0.98, samples=2] {20/17};
\addplot[black!12, forget plot] fill between[of=sv and cs];
% J^o - J_0 = minimum of the two curves (thick, under the curves)
\addplot[black!30, line width=2.6pt, domain=0.5:0.71667, samples=60] {0.5*(x/(1-x)-3/17)};
\addlegendentry{$\Jopt-\bar J=\min\{\text{solid},\text{dashed}\}$}
\addplot[black!30, line width=2.6pt, forget plot, domain=0.71667:0.98, samples=2] {20/17};
\addplot[black, line width=0.7pt, domain=0.5:0.98, samples=120] {0.5*(x/(1-x)-3/17)};
\addlegendentry{saved between the checks $=J_{\SK}-\bar J$}
\addplot[black!60, dashed, line width=0.7pt, domain=0.5:0.98, samples=2] {20/17};
\addlegendentry{cost before either check}
\addplot[black, only marks, mark=*, mark size=1.4pt, forget plot] coordinates {(0.9,4.412) (0.98,24.412)};
\addplot[black, only marks, mark=o, mark size=1.4pt, forget plot] coordinates {(0.9,1.1765) (0.98,1.1765)};
\node[font=\scriptsize] at (axis cs:0.966,5.48) {$\Delta J$};
\draw[black!70, densely dashdotted] (axis cs:0.71667,0.2) -- (axis cs:0.71667,1.1765);
\node[font=\scriptsize, anchor=south west] at (axis cs:0.726,0.235) {$h^\ast$};
\end{semilogyaxis}
\end{tikzpicture}
\caption{Sequencing the fork of the release net. The supervisor $\Sopt$ refuses $\tr{ci}$ before the scan if and only if the solid curve lies above the dashed line, i.e., $h>h^\ast=43/60$. The thick curve $\Jopt-\bar J$ is the smaller of the two, and the shaded gap is $\Delta J$ in Eq.~\eqref{eq:gap}. Filled and open dots are the values of $J_{\SK}-\bar J$ and $\Jopt-\bar J$, respectively, in the ``after tests'' rows of Table~\ref{tab:intrusive}.}
\label{fig:fork}

\end{figure}
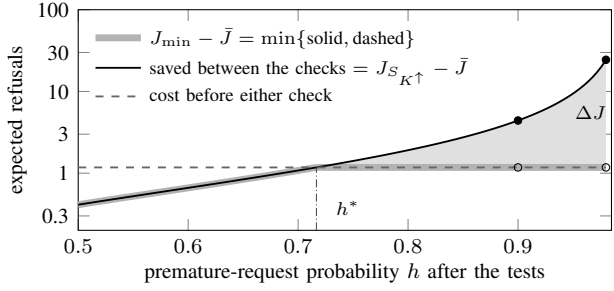

We now quantify the effect of this decision. The probability that $\SK$ runs $\tr{ci}$ first is $w=\frac12$. Refusing $\tr{ci}$ before the scan costs $w$ times the right-hand side of Eq.~\eqref{eq:fork}, i.e., $20/17$ refusals before either check. In return, it saves $w$ times the left-hand side, i.e., $\frac12\big(h/(1-h)-3/17\big)$, between the checks. Fig.~\ref{fig:fork} plots both quantities against $h$. The solid curve shows the refusals saved between the checks, while the dashed line shows the cost before them. Since $\tr{fork}$ fires once, the $\Delta J$ of Table~\ref{tab:intrusive} for premature requests after the tests only is $\max\{0,w\delta\}$, i.e.,
\begin{equation}\label{eq:gap}
J_{\SK}(M_0)-\Jopt(M_0)=\max\big\{0,\ \tfrac{h}{2(1-h)}-\tfrac{43}{34}\big\}.
\end{equation}
Thus, the maximally permissive least-intrusive supervisor $\Sopt$ takes into account where the proposer makes premature requests, here at $M_{\{1\}}$, while the conventional maximally permissive supervisor $\SK$ is defined without the proposer model.

\begin{remark}
We remark that, since the terminal cost is zero on $\F$, $J$ counts refusals and calls only. A terminal cost on $\pl{abort}$, which only the uncontrollable $\tr{fail}$ and $\tr{reject}$ mark, changes neither $\Sopt$ nor Table~\ref{tab:intrusive}. However, a terminal cost $g_{\mathrm a}$ on the markings with $\pl{aborted}$ marked has an effect only beyond a threshold. Specifically, it changes neither $\Sopt$ nor Table~\ref{tab:intrusive}, for every $\ca$, if and only if $g_{\mathrm a}\le g_{\mathrm a}^\ast:=1+5507/1037\approx6.31$. The term $5507/1037$ is the expected number of refusals until a rollout caught by a freeze goes live without $\tr{abort}$. For $g_{\mathrm a}>g_{\mathrm a}^\ast$, $\Sopt$ also refuses $\tr{abort}$. Even then, $h^\ast$ and the monitor place remain unchanged, while every $\Delta J$ increases by $\tfrac{306}{52175}(g_{\mathrm a}-g_{\mathrm a}^\ast)$. The factor $306/52175\approx0.006$ is the probability that a run under $\SK$ ends in $\pl{aborted}$.
\end{remark}

\subsection{Prompt injection}\label{sec:inject}

\begin{figure}[!t]
\centering
\includegraphics[width=\columnwidth]{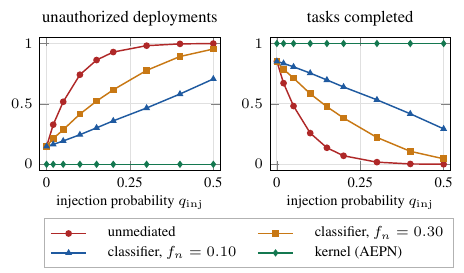}
\caption{Injection probability $q_{\mathrm{inj}}$ swept from $0$ to $0.5$ on the release net under a scripted proposer, with $10^4$ episodes per point. The left panel shows the fraction of episodes with an unauthorized deployment, i.e., a $\tr{start}$ executed without an approval, injected or not. The right panel shows the fraction of episodes that complete the task. The classifier curves use the false-negative rates $f_n=0.10$ and $f_n=0.30$.}
\label{fig:inject}
\end{figure}

Finally, we test the admission rule against prompt injection. To this end, Fig.~\ref{fig:inject} compares three architectures under a scripted proposer. The proposer emits the correct next action, except that an injected instruction makes it propose $\tr{start}$ immediately with probability $q_{\mathrm{inj}}$, and that it proposes a legitimate action two steps early with probability $0.15$.

The script has no separate environment model and instead emits every step of the workflow in order, thereby also acting as the environment and the reviewer. An early proposal is the action two positions ahead in this order, and the injection is drawn before it. The outcomes $\tr{ok}$, $\tr{approve}$ and $\tr{finish}$ always succeed, and no incident occurs. For simplicity, each emitted step, outcomes included, counts as one proposal. Since an early outcome names a transition of another principal, it would map to $\bot$ in Eq.~\eqref{eq:admit}. Hence, such an outcome is refused.

We now describe the three architectures. The first one, \emph{unmediated}, executes every proposal. In this architecture, an early benign proposal that the net does not yet enable fails at its tool without effect. The exception is $\tr{start}$. A $\tr{start}$ executed without an approval, whether injected or proposed early in place of $\tr{request}$, is an unauthorized deployment. Such a deployment ends the episode, which then counts as not completed. The second architecture, \emph{classifier}, blocks each injected proposal independently with probability $1-f_n$, where $f_n$ is its false-negative rate, and it never blocks a benign one. The third architecture, \emph{kernel}, applies the admission rule Eq.~\eqref{eq:admit} without any detection. For all three architectures, each episode has at most 60 proposals, and the standard error is at most $0.005$.

As shown in Fig.~\ref{fig:inject}, a classifier only lowers the curve of unauthorized deployments. Since it cannot block the benign but premature $\tr{start}$, its curves start at about $0.15$ at $q_{\mathrm{inj}}=0$, like the unmediated one. In contrast, the kernel curve stays at zero for every phrasing of the injected instruction and every supervisor $S$ in Eq.~\eqref{eq:admit}, since $\tr{start}$ is not enabled while $\pl{approved}$ is empty.

In addition, the kernel completes every task of the script, since a refused premature action leaves the workflow intact and the script then proposes the next step. The cost is about $1.5$ refusals of benign proposals per episode at every $q_{\mathrm{inj}}$, each naming an action that the net does not yet enable. Refused injections add to this count. An LLM agent, unlike the script, may abandon the task after a refusal.

Consider now an injection aimed at the supervisor in check~iv) of Eq.~\eqref{eq:admit}, rather than at the enabling condition in check~iii). For instance, an injected ``publish now'' that arrives while a staged release is frozen maps to $\tr{publish\_fr}$, which the net enables. Nevertheless, the kernel will refuse $\tr{publish\_fr}$, since neither $\SK$ nor its refinement $\Sopt$ admits a transition into $\B$. This transition is the only enabled one that $\SK$ refuses on this net. The injection in the figure, however, proposes only $\tr{start}$ and thus tests the enabling condition alone. Since the proposer is a script, the figure shows what each architecture does once an LLM is hijacked, but it does not show how often this happens.

\section{Conclusion}\label{sec:concl}

In this paper, we have formalized the execution harness of LLM agents as a supervisory control problem on a Petri net and shown that a supervisor that is safe on the net keeps the running system safe under four assumptions. Moreover, we have introduced the replanning closed loop, in which a refused agent may propose again at a cost. In this loop, the conventional maximally permissive supervisor can be arbitrarily more intrusive than a stricter safe one. For a given proposer model, the maximally permissive least-intrusive supervisor is characterized as a value-threshold refinement of the conventional one. Under precise conditions, this refinement orders two concurrent actions by at most one monitor place. On the release agent, we have found by synthesis that safe deployment requires a step that the supervisor can refuse after a rollout starts. In future work, we aim to bound the extra cost of an inaccurate proposer model and to loosen our assumption that it depends only on the current marking. Moreover, we plan to treat partial observation of the marking and to develop compositional synthesis for large nets.

\bibliographystyle{IEEEtran}
\bibliography{references_final,references_ctl}

\end{document}